\pdfoutput=1

\documentclass[11pt]{article}

\usepackage[final]{acl}

\usepackage{times}
\usepackage{latexsym}
\usepackage{graphicx}
 \usepackage{array,multirow}
 \usepackage{booktabs}

 \usepackage{algorithm}
\usepackage{algpseudocode}

\algtext*{EndWhile}
\algtext*{EndIf}
\algtext*{EndFor}

\usepackage[T1]{fontenc}

\usepackage[utf8]{inputenc}

\usepackage{microtype}

\usepackage{inconsolata}

\usepackage{graphicx}
\usepackage[most]{tcolorbox}
\usepackage{adjustbox}
\usepackage{makecell}

\newcommand{\Judge}{\textit{LLM-as-a-qualitative-judge}}

\title{LLM-as-a-qualitative-judge:\\automating error analysis in natural language generation}

\author{
Nadezhda Chirkova$^1\quad$
\textbf{Tunde Oluwaseyi Ajayi}$^2\quad$ 
\textbf{Seth Aycock}$^3\quad$
\textbf{Zain Muhammad Mujahid}$^4\quad$\\
\textbf{Vladana Perlić}$^{5,6}\quad$
\textbf{Ekaterina Borisova}$^{7,8}$
\textbf{Markarit Vartampetian}$^9\quad$
   \\ \\
   $^1$Naver Labs Europe \\
   $^2$Insight Research Ireland Centre for Data Analytics,  Data Science Institute,  University of Galway\\
   $^3$University of Amsterdam
   $^4$University of Copenhagen
   $^5$STMicroelectronics
   $^6$Télécom Paris \\
   $^7$Deutsches Forschungszentrum für Künstliche Intelligenz GmbH (DFKI)\\
   $^8$Technische Universität Berlin\\
   $^9$Univ. Grenoble Alpes, CNRS, Grenoble INP, LIG, 38000 Grenoble, France\\
   \textbf{Correspondence}: \texttt{nadia.chirkova@naverlabs.com}
        \\
}

\begin{document}
\maketitle
\begin{abstract}
Prompting large language models (LLMs) to evaluate generated text, known as \textit{LLM-as-a-judge}, has become a standard evaluation approach in natural language generation (NLG), but is primarily used as a \textit{quantitative} tool, i.e. with numerical scores as main outputs. In this work, we propose \textit{LLM-as-a-qualitative-judge}, an LLM-based evaluation approach with the main output being a \textit{structured report} of \textit{common issue types} in the NLG system outputs. Our approach is targeted at providing developers with meaningful insights on what improvements can be done to a given NLG system and consists of two main steps, namely open-ended per-instance issue analysis and clustering of the discovered issues using an intuitive cumulative algorithm. We also introduce a strategy for evaluating the proposed approach, coupled with $\sim$300 annotations of issues in instances from 12 NLG datasets. Our results show that instance-specific issues output by \textit{LLM-as-a-qualitative-judge} match those annotated by humans in 2/3 cases, and that LLM-as-a-qualitative-judge is capable of producing error type reports resembling the reports composed by human annotators. We also demonstrate in a case study how the use of LLM-as-a-qualitative-judge can substantially improve NLG systems performance. Our code and data are publicly available\footnote{Code \& data: \url{https://github.com/tunde-ajayi/llm-as-a-qualitative-judge}}. 
\end{abstract}

\section{Introduction}
Prompting large language models (LLMs) to output evaluation scores, known as \textit{LLM-as-a-judge}~\citep{zheng23-judgingllmasajudgemtbench},
has become a standard approach for evaluating performance in natural language generation (NLG) tasks. In contrast to classic statistical measures such as BLEU \citep{papineni-etal-2002-bleu}, ROUGE \citep{lin-2004-rouge}, or METEOR \citep{banerjee-lavie-2005-meteor}, which primarily rely on surface-level lexical overlap, LLM-as-a-judge evaluates text based on deep semantic understanding, allowing it to better handle diverse phrasings that convey equivalent meanings. Consequently, it shows stronger alignment with human judgment in various tasks, including machine translation \citep{kocmi-federmann-2023-large}, summarization~\citep{seahorse}, or open-ended instruction following~\citep{flask}, especially with strong recent LLMs. 

\begin{figure}[t!]
     \centering
     \includegraphics[width=\linewidth]{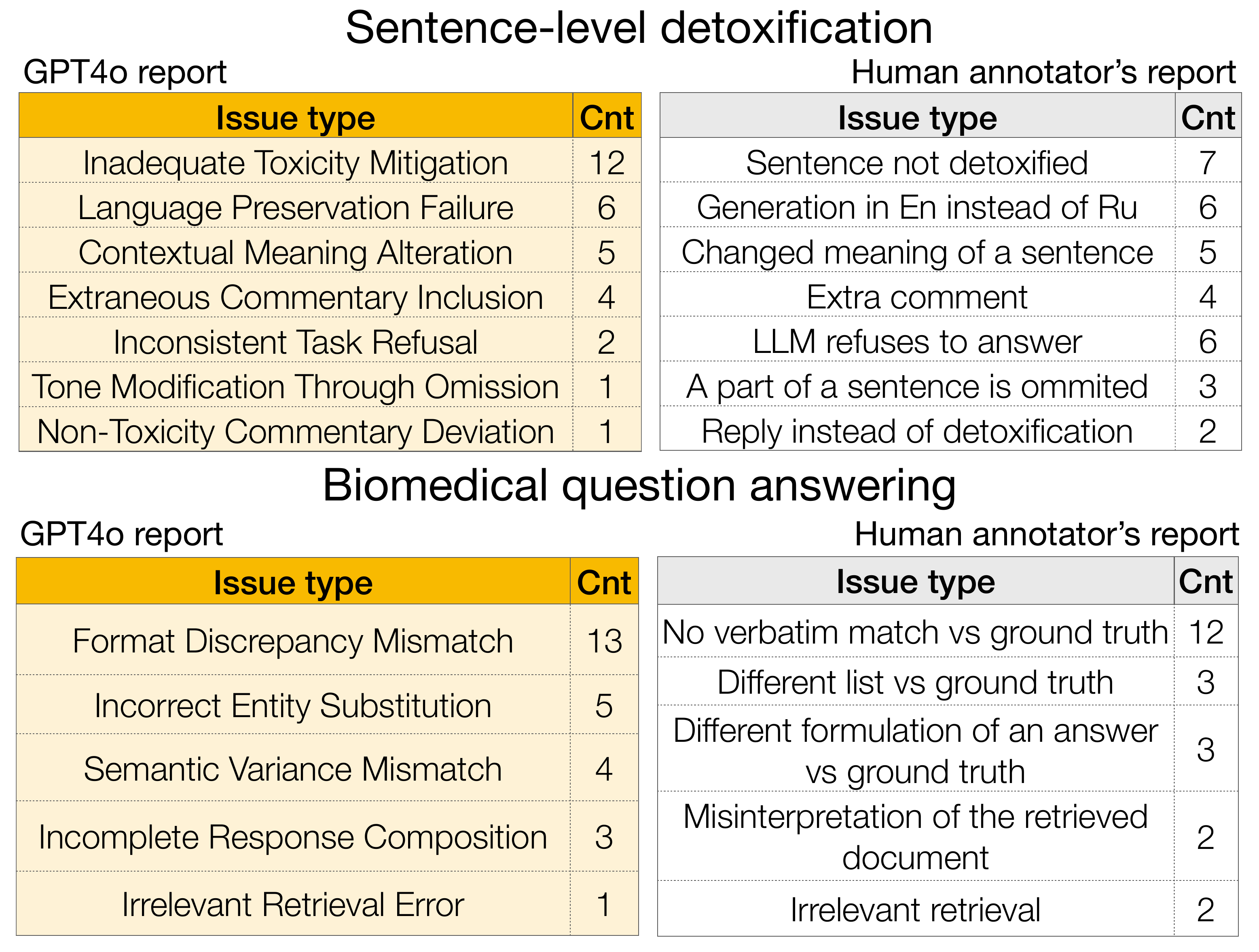}
     \caption{Issue types reports for two datasets composed by the proposed \textit{LLM-as-a-qualitative-judge} (GPT-4o) and by a human annotator. All steps of analysis performed by GPT-4o, including error types formulation and error grouping. The full generated report also includes comprehensive error type descriptions, omitted here due to the space limit. Cnt represents issue type counts. 
     }
     \label{fig:summary_examples}
 \end{figure}

 \begin{figure*}[t] 
    \center{\includegraphics[width=1.0\textwidth]{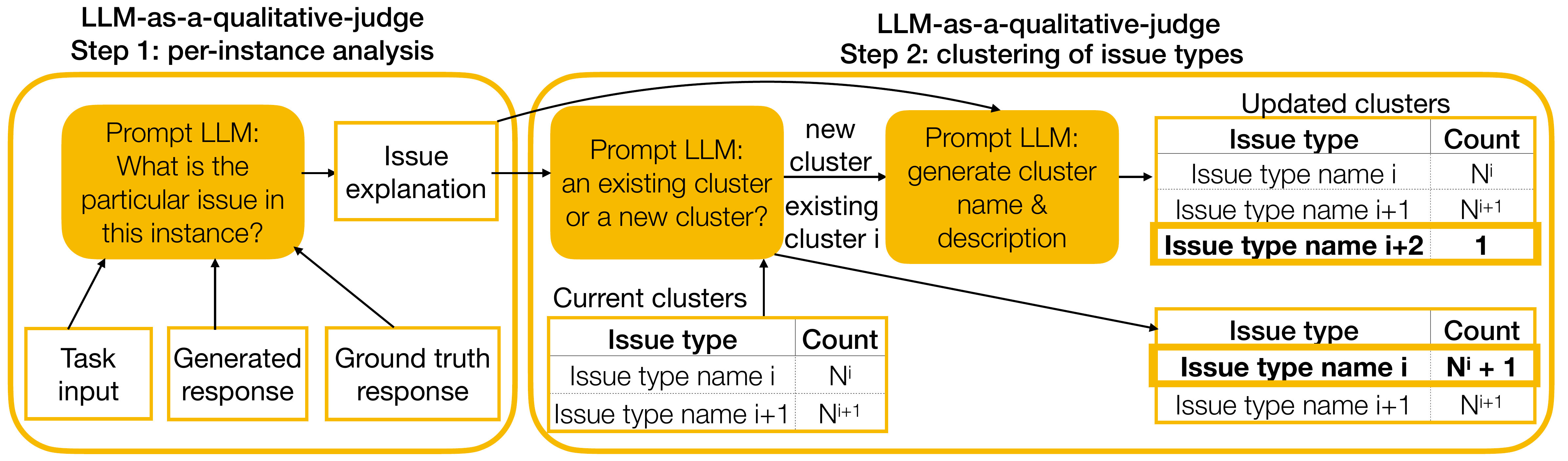}}
    \caption{
    Illustration of the proposed LLM-as-a-qualitative-judge approach.
    }
    \label{fig:illustration}
\end{figure*}

Recent works propose various extensions of the LLM-as-a-judge approach, including pairwise model comparison~\citep{zheng2023judging}, finetuning LLMs for evaluation~\citep{prometheus,prometheus2}, multi-criteria evaluation~\citep{liang2023holistic,fu-etal-2024-gptscore}, the dynamic selection of evaluation criteria~\citep{flask}, or even using per-instance evaluation checklists~\citep{checklists,biggen}. 
A common technique to improve LLM-as-a-judge is to 
ask a model to output an explanation for the predicted score(s).

However, even with the extensions listed above, LLM-as-a-judge remains primarily a \textit{quantitative} evaluation tool, i.e., the final result used by researchers and practitioners is \textit{quantitative evaluation scores}. 
At the same time, language generation is a complex multi-faceted task with a vast space of potential issues, including in various aspects of generated texts (grammaticality, factuality, logical coherence, etc.), in preprocessing of the input data and postprocessing of the NLG outputs, or even with user requests. An effective and commonly used strategy for spotting such issues is a manual \textit{qualitative error analysis} of a subset of predictions, 
which allows developers to identify artifacts, fix system issues, and detect flaws in quantitative evaluation. 
Yet this analysis is often skipped in practice~\cite{van-miltenburg-etal-2023-barriers, van-miltenburg-etal-2021-underreporting}, due to overreliance of developers on quantitative metrics, as well as high demand in terms of time and effort needed to conduct such analysis.

In this work, we introduce \Judge, a novel approach which automates error analysis, with the main output being a
\textit{a structured report} aggregating the common \textit{qualitative} error types in the NLG outputs for a given dataset. 
The two key steps of LLM-as-a-qualitative-judge are (1) open-ended per-instance error analysis and (2) clustering of the discovered error types. Per-instance analysis implies prompting an LLM to detect an issue in the given NLG system output, where an issue may be arbitrary, i.e. we do not provide any predefined set of possible issues. For error clustering, we propose an intuitive and effective algorithm which resembles how humans solve the corresponding task. An example of the generated report is presented in Figure~\ref{fig:illustration} (left), and a high-level illustration of the proposed approach is presented in Figure~\ref{fig:illustration} (right).

To summarize, our contributions are three-fold:

\begin{itemize}
    \item We introduce an \Judge, a novel approach for LLM-based evaluation, outputting a structured report of common error types in a dataset;
    \item In a case study on BigBenchHard tasks, we demonstrate that LLM-as-a-qualitative-judge can substantially improve the performance of NLG systems;
    \item We collect $\sim$300 manual annotations of open-ended issues in the instances coming from 12 diverse NLG datasets, as well as the manual annotations of their per-dataset clustering;
    \item We introduce a strategy for meta-evaluating \Judge~and show that \Judge~is capable of producing error type reports which resemble the reports produced by humans.
\end{itemize}

We hope that the proposed \Judge~approach will reduce the time and effort required for issue analysis and will help 
practitioners to more easily improve their NLG pipelines. Our code and data are available as \url{https://github.com/tunde-ajayi/llm-as-a-qualitative-judge}.

\section{Proposed approach}
\label{sec:method}
The main goal of our proposed approach, \Judge, is to provide a developer with a \textit{structured report} of the main \textit{types of issues} (and their counts) in the outputs of a given NLG system for a given dataset. 
In the rest of the work, we use terms \textit{issues}, \textit{errors}, or \textit{failures} interchangeably to denote any problems which may occur in the NLG outputs. 
Examples of such problems include (but are not limited to) unfinished generation due to reaching the maximum new tokens limit, an error in one of the reasoning steps, a problem with the retrieved documents in retrieval-augmented generation, or an error in evaluation due to the use of an inappropriate metric.  
We do not employ any predefined set of possible issues, and use the term \textit{open-ended issue analysis} to refer to the problem of detecting arbitrary issues in NLG outputs.

For the purposes of our algorithm, the dataset consists of \textit{instances}, each represented by a \textit{task input} (a string containing a task instruction and the input data), a \textit{ground truth response} (a string defining a correct answer), and a \textit{generated response} (a final output of the NLG system). Each instance can be optionally augmented with other fields, e.g., the intermediate outputs of an NLG system such as retrieved documents in retrieval-augmented generation, or additional information on the NLG system, e.g., a definition of a task metric. 
The \Judge~algorithm is summarized in Algorithm~\ref{alg:proposed}, illustrated in Figure~\ref{fig:illustration} and described step-by-step below.

\paragraph{Preliminary step: detecting examples with errors.} 
Our algorithm focuses only on the instances from the dataset with any sort of issues. We rely on the task-specific quantitative metric to select such instances, i.e., instances which did not get high scores in the quantitative evaluation.

\paragraph{Step 1: per-instance analysis.}
For each instance, we prompt an LLM to identify \textit{``one, most important, specific, clearly visible issue''}, provided with a task input, a ground truth response, a generated response, and optionally other fields as described above. We prompt an LLM to output a detailed analysis of a given instance, followed by a special separator and a final 1--2 sentence description of an identified issue, which is referred to as a \textit{per-instance issue explanation} in the following steps of the approach.
The particular prompt used for per-instance analysis is presented in App. Figure~\ref{fig:per_ex_analysis_prompt}.

\begin{algorithm}[t]
\caption{LLM-as-a-qualitative-judge}\label{alg:proposed}
\begin{small}
\begin{algorithmic}[1]
\Require a list of task inputs $U$, a list of ground truth responses $R^{\mathrm{gt}}$, a list of generated responses $R$ --- all of length $N$; 
\Ensure a report $C$ listing issue types and their counts; \newline a list  $A$ of per-instance issue explanations
\State $A \gets []$ // \textit{empty initialization for per-instance analysis}
\State $C \gets []$ // \textit{empty initialization for a report}
\For{$i = 1, \dots, N$}
    \State // \textit{per-instance analysis}
    \State $A[i] \gets \mathrm{LLM}(\mathrm{prompt}_\mathrm{analysis}; U[i], R^{\mathrm{gt}}[i], R[i])$ 
     \State // $A[i]$ \textit{is a string containing issue explanation}
     \State // \textit{report generation}
    \If {$i > 1$}
    \State // \textit{an existing issue type or a new one?}
    \State $K \gets \mathrm{LLM}(\mathrm{prompt}_\mathrm{decision}; A[i], C)$ 
    \State // $K \in \{1,\dots, |C|, \mathrm{None}\}$
    \Else 
    \State $K \gets \mathrm{None}$ //  \textit{first step is always new issue type}
    \EndIf
     \If {$K$ is None}
      \State // \textit{create a new issue type}
     \State $E \gets \mathrm{LLM}(\mathrm{prompt}_\mathrm{new\_type}; A[i]) $
     \State // $E$ \textit{is a dictionary containing a short issue name and an issue description}
     \State $E[\mathrm{``count"}] \gets 1$
     \State $C.\mathrm{append}(E)$
     \Else
     \State // \textit{augment an existing issue type}
     \State $C[K][\mathrm{``count"}] \gets C[K][\mathrm{``count"}] + 1$
    \EndIf
\EndFor
\State \textbf{return} $C$, $A$
\end{algorithmic}
\end{small}
\end{algorithm}

\begin{figure*}[t] 
    \center{\includegraphics[width=1.0\textwidth]{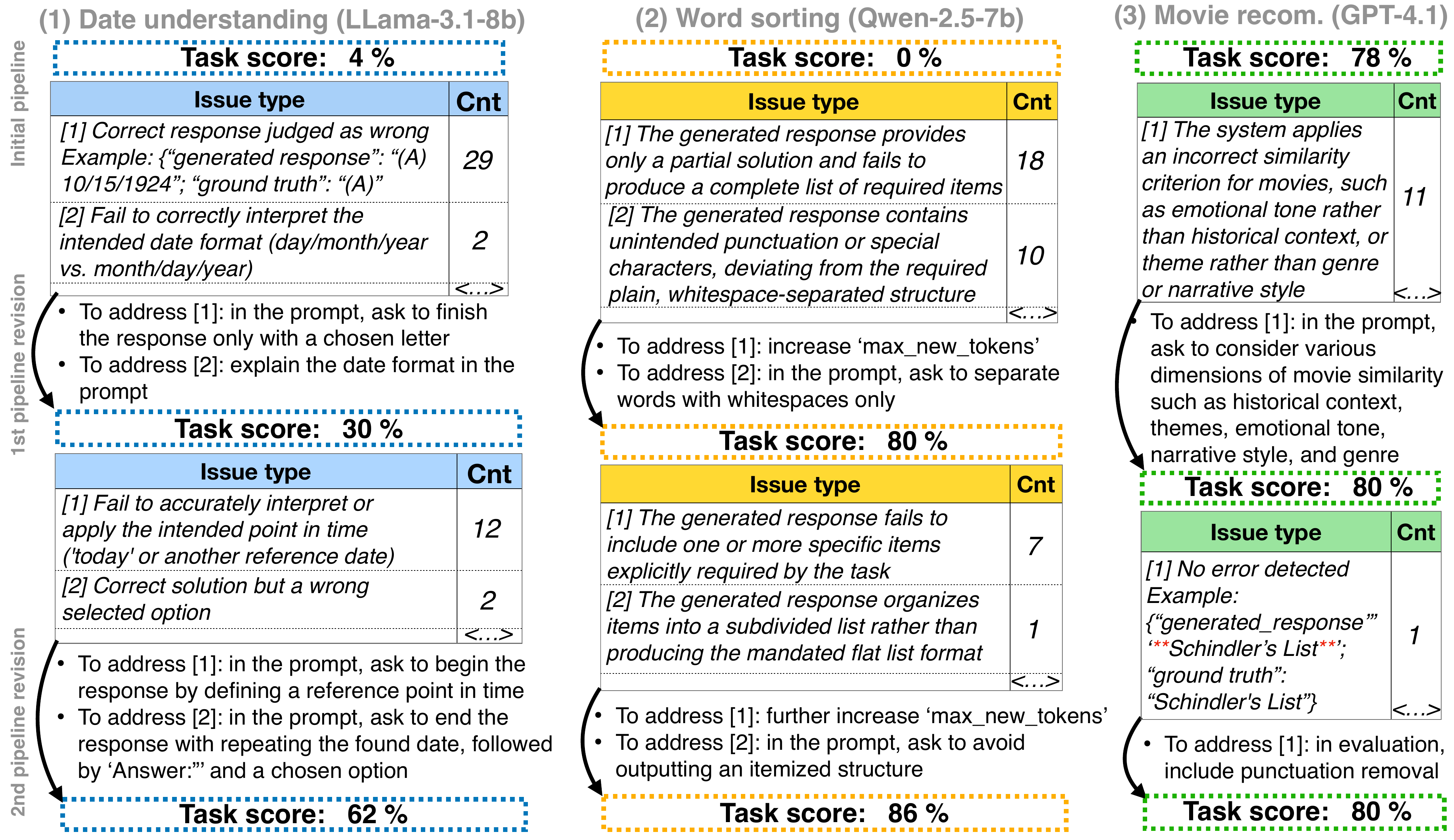}}
    \caption{Case study on three BigBenchHard tasks: after building a simple pipeline for a task, we perform two rounds of generating issue reports with LLM-as-a-qualitative-judge (a table with issue types and their counts) and manually revising the pipeline based solely on the generated reports. Task performance is improved in all cases.}
    \label{fig:case}
\end{figure*}

\paragraph{Step 2: issue clustering.}
The second step in \Judge~consists of clustering issues discovered in the first step and forming a final report of main issue types based on the clustering results. This can be, in principle, done with any clustering algorithm, e.g., k-means with BERT-based embeddings~\citep{bert} or \textit{directly prompting} a strong LLM to output clustering~\citep{DBLP:journals/tacl/0002GGLWN24}, provided with clustering inputs in a single prompt. In the experiments, we demonstrate the downsides of these approaches, e.g., 
classic approaches perform poorly on our data, and
clustering with direct prompting fails for larger datasets, weaker LLMs, and does not ensure the structural correctness of the generated report.

Inspired by how humans would cluster issues, we propose an intuitive \textit{cumulative} issue clustering algorithm. Our clustering algorithm goes through instances one-by-one and gradually builds the issue types report. For each instance, we provide an LLM with the current report and the per-instance issue explanation, and prompt the LLM to \textit{decide} whether this issue explanation can be attributed to one of the already discovered issue types (clusters) or it should form a new issue type (cluster). In the former case, we augment the counter of the corresponding issue type by one. In the latter case, we also prompt an LLM to formulate a short name and a 1--2 sentence description of a new error type, based on the per-instance issue description. In particular, we instruct an LLM to formulate \textit{``a fine-grained issue type that can be generalized to other instances''}. The new issue type is then added to the report, represented by the generated issue type name, description, and the counter set to one. The first instance in the dataset is always a new issue type. 
Appendix Figures~\ref{fig:summary_cum_strategy_prompt} and~\ref{fig:summary_rag_cumulative_newtype_prompt} show the prompts used for the two described clustering steps.

The final issue types report is composed of issue type names and descriptions, paired with the counts of how many instances were attributed to the corresponding issue type.

\section{Case study}
\label{sec:case}
Our first set of experiments is targeted at demonstrating a practical utility of the proposed \Judge.

\paragraph{Experimental setup.} We pick three tasks from a BigBenchHard collection~\citep{bbh}, namely Date understanding, Word sorting, and Movie recommendation. For each dataset, we build a simplest pipeline consisting of prepending a simple system prompt ``\textit{You are a helpful assistant. Output your answer after a final separator `Answer:'}'', LLM generation with default hyperparameters, and a string match-based evaluation function. We then perform two rounds of generating an issue report with \Judge~(GPT-4.1) and manually revising the pipeline solely based on the generated report (issue types, their counts, and possibly 1 example of each issue type). More details are given in Appendix~\ref{sec:expsetup}.

\paragraph{Results.} As shown in Figure~\ref{fig:case}, the task performance is improved in all three cases.
For example, in the Date understanding task, revisions inspired by the generated issue reports include explaining the date format in the prompt, suggesting to begin the response with determining a reference point in time, and providing a specific template for the output. These revisions improved performance from 4\% to 62\%.

\section{Meta-evaluation methodology}
\label{sec:evaluation}
This section described a methodology that we propose to meta-evaluate \Judge. The corresponding set of experiments aims both to assess the effectiveness of two steps and to identify the optimal configurations for \Judge.

\paragraph{Real-world data.} We manually annotate per-instance issues and their per-dataset clustering for a diverse pool of 12 datasets, with various open-source LLMs as generators. We consider 7 generative tasks, and for one of the tasks, namely retrieval-augmented question answering (RA-QA), we consider 6 domains. All the labeling was performed by the authors of the paper. The final dataset comprises 297 instances. Table~\ref{tab:data} provides the data summary. More details on data annotation are presented in Appendix~\ref{sec:annotation}.

\paragraph{Synthetic data.} We also consider synthetic data for evaluating clustering: we define a set of possible issue types $e$ and their frequencies $n_e$, then prompt GPT-4o to reformulate each issue $e$ in various ways $n_e$ times, and then use this data as per-instance analysis for clustering. This allows us to evaluate clustering on larger datasets, i.e., 100-1000 instances.

\paragraph{Metrics.} For per-instance analysis, we prompt an \textit{evaluator LLM} to judge whether the issue explanation determined by the \Judge~for a particular instance matches the issue determined by the human annotator. The outputs from the \textit{evaluator LLM} are binary and are accumulated into a \textit{per-instance analysis accuracy} score.

\begin{table}[t!]
  \centering
  \scriptsize
  \begin{tabular}{p{2.2cm}p{3.5cm}p{0.5cm}}
    \toprule
    \textbf{Task} & \textbf{Dataset reference} & \textbf{\# ex.} \\
    \midrule
    \multicolumn{3}{l}{\textbf{Natural Language Generation}}  \\ \midrule
    Instruction following & FLASK \cite{flask} & 34\\  
    Translation en-ru & WMT'22~\cite{kocmi-etal-2022-findings} & 38\\  
    Long context QA & Elitr-Bench \cite{DBLP:conf/coling/ThonetBR25} & 26\\  
    Semantic parsing & PIZZA \cite{DBLP:journals/corr/abs-2212-00265} & 34\\
    Grade school math & GSM8K \cite{DBLP:journals/corr/abs-2110-14168} & 17\\  
    Detoxification & ParaDetox \cite{dementieva2024overview} & 36\\ \midrule   
    \multicolumn{3}{l}{\textbf{Retrieval-augmented QA}}  \\ \midrule
    Factoid QA in Russian  & MKQA (ru) \cite{longpre-etal-2021-mkqa} & 29\\ 
    Biomedical QA  & BioASQ \cite{krithara2023bioasq} & 27\\  
    Lifestyle forum QA & \multirow{2}{3.5cm}{RobustQA \cite{DBLP:conf/emnlp/Han00XWLWMC24,lotte}} & 21\\  
    Search engine queries  & SearchQA \cite{DBLP:journals/corr/DunnSHGCC17} & 13\\  
    Educational QA & SyllabusQA \citep{DBLP:conf/acl/FernandezSL24} & 9 \\ 
    \midrule
    Total & & 297 \\ 
    \bottomrule
  \end{tabular}
  \caption{The statistics of the annotated evaluation data. 
  }
  \label{tab:data}
\end{table}

We evaluate cluster agreement using a \textit{Rand index adjusted for chance}, or Adjusted Rand Index (ARI)\footnote{We use the scikit learn implementation: \url{https://scikit-learn.org/stable/modules/generated/sklearn.metrics.adjusted_rand_score.html}.}. We also evaluate the agreement in error type descriptions, by finding the best possible mapping between clusters found by a human annotator and by \Judge, and then prompting an \textit{evaluator LLM} to judge the semantic equivalence of the corresponding issue type descriptions. This metric is denoted as Semantic Label Consistency (SLC).

\section{Meta-evaluation experiments}
\label{sec:results}
\subsection{Experimental setup}
We test per-instance analysis with a range of commercial and open-source LLMs, and issue clustering with three LLMs: GPT-4o, Gemini-2-Flash, and Qwen-2.5-7B.
For issue clustering, we compare the proposed cumulative clustering approach to the direct LLM prompting and classic clustering approaches. All clustering runs operate on the per-instance analysis output by GPT-4o, and clustering results are averaged over 3 runs.

Tables~\ref{tab:llm_info} and \ref{tab:dataset_licenses} in Appendix list references and licenses of the used models and datasets, respectively. Prompts used for all the stages 
are presented in the Appendix~\ref{sec:prompts}. Exact task formulations in prompts were adjusted by only using three RA-QA datasets (MKQA (ru), RobustQA Lifestyle and Writing).

For classic clustering approaches, we use the \verb|scikit-learn| implementation with BERT embeddings and tune hyperparameters as described in App.~\ref{sec:clustering-setup}. For methods requiring the number of clusters, we set it the same as in the annotator's data.

More details on experimental settings and as well as results on meta-evaluating the evaluator LLM are presented in Appendix~\ref{sec:expsetup}.

Appendix~\ref{app:examples} presents the examples of generated error type reports, per-instance issue explanations, and confusion matrices for all considered datasets. We also provide an archive with experimental results in the project repository\footnote{A zip archive with experimental results is available at \url{https://github.com/tunde-ajayi/llm-as-a-qualitative-judge/tree/main/data}}.

\subsection{Per-instance analysis}

Table~\ref{tab:per_example_analysis} reports the performance of various LLMs in per-instance analysis. Strongest LLMs, including commercial LLMs and a larger open-source Qwen-2.5-32B achieve an accuracy of 62--67\%, i.e. about 2/3 of issues in our dataset were successfully correctly explained by strong models. The accuracy of open-source LLMs is substantially influenced by their size: Qwen-2.5 accuracy raises from 32\% to 67\% when increasing size from 1.5B to 32B. Various LLMs of 7--8B size demonstrate analysis accuracy of 42--60\%. 

We note that our results are consistent with previously reported findings in the literature regarding the typical level of agreement between LLM-based evaluations and human judgments. For example, FLASK reports the highest  correlation between model-based evaluation and human labelers, of 68\% (Table 1 in \citealp{flask}), or METAL reports the highest agreement between the LLM evaluators and human scores of 59-82\% for the first three criteria in (\citealp{metal}, Table 3, English).

\textit{To sum up, our results demonstrate the high effectiveness of strong LLMs in open-ended issue explanation for generative tasks. For practical applications, we recommend using recent models such as GPT-4o, Gemini-2.5-Flash or Qwen-2.5-32B.}

\begin{table}[t!]
  \centering
  \small
  \begin{tabular}{p{4.5cm}r}
    \toprule
    \textbf{Model} & \textbf{Accuracy (\%)} \\
    \midrule
    GPT-4o & 66.3 \\
    Gemini-2.0-Flash & 65.0 \\
    Qwen-2.5-32B & 68.7 \\
    Qwen-2.5-7B & 55.5 \\
    Qwen-2.5-1.5B & 30.7 \\
    DeepSeek-R1-Distill-Llama-8B & 56.1 \\
    Aya-Expanse-8B & 42.1 \\
    Llama-3.1-8B-Instruct & 55.4 \\
    Ministral-8B-Instruct-2410 & 58.1 \\
    \bottomrule
  \end{tabular}
  \caption{Performance of various LLMs in per-instance analysis. Evaluator LLM: Claude-3-7-Sonnet-20250219.}
  \label{tab:per_example_analysis}
\end{table}

\begin{figure*}[ht] 
    \center{\includegraphics[width=\textwidth]{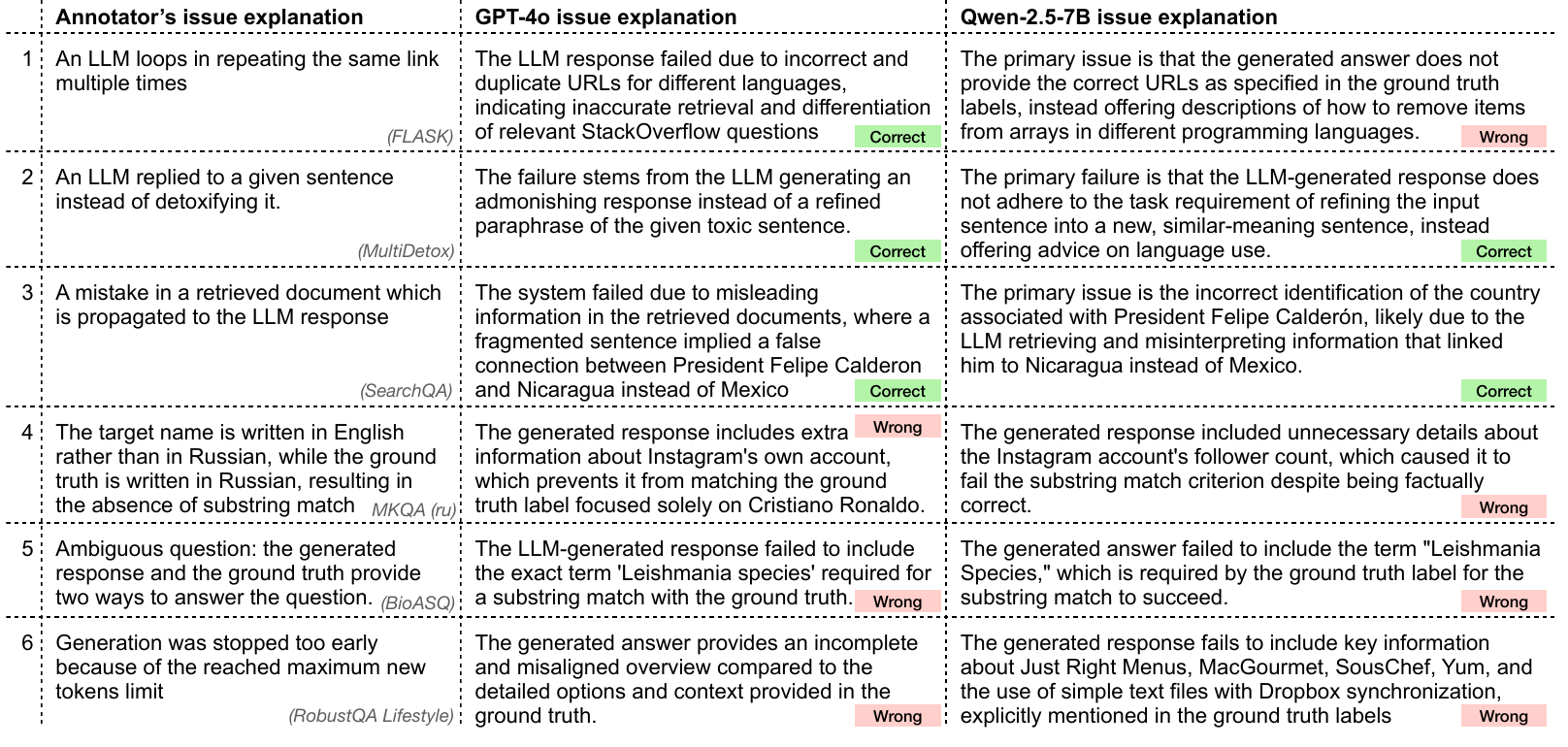}}
    \caption{Examples of per-instance analysis.}
    \label{fig:eval_examples}
\end{figure*}

\begin{figure*}[ht!] 
    \includegraphics[width=\linewidth]{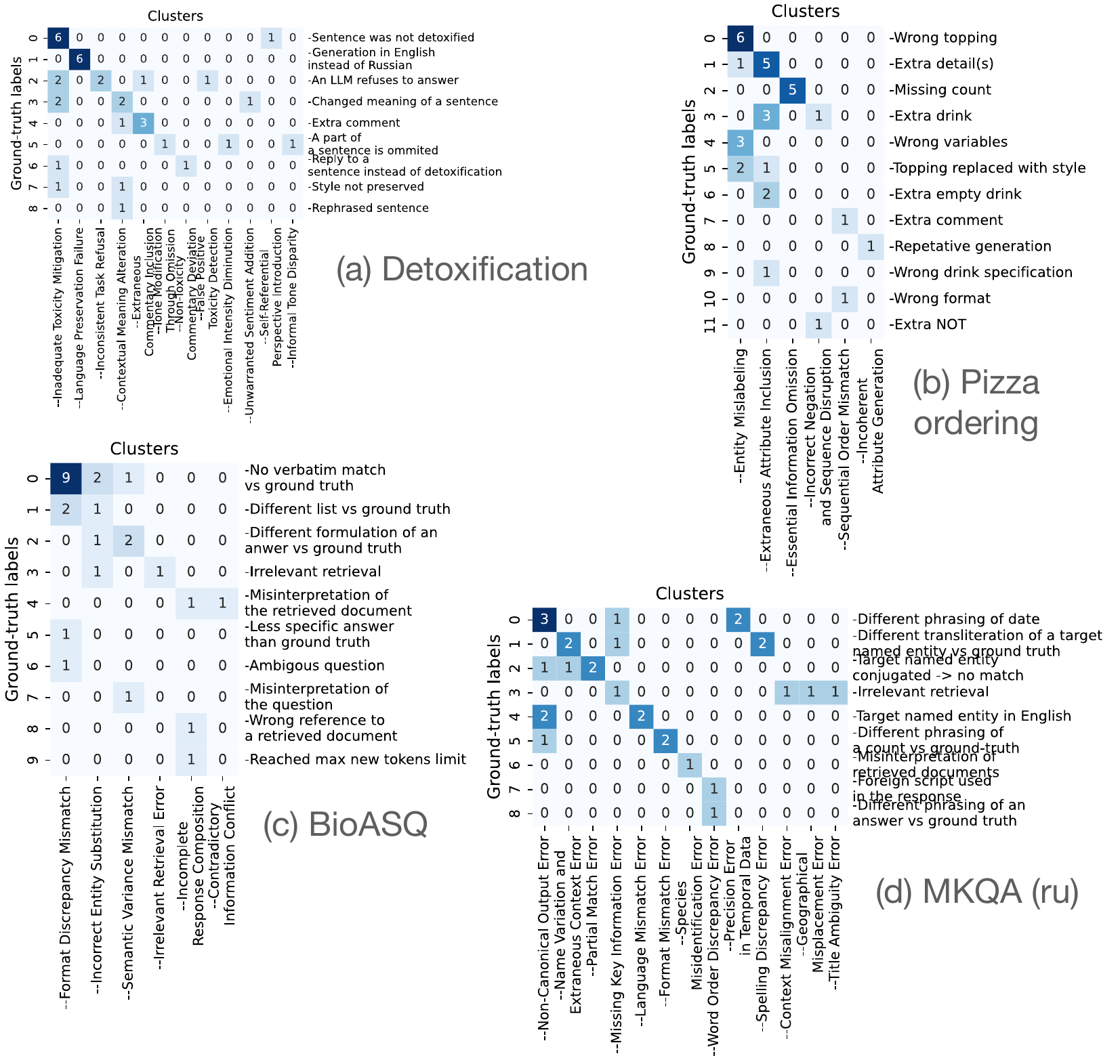}
    \caption{Examples of confusion matrices visualizing clustering agreement between \textit{LLM-as-qualitative-judge}-generated and the annotator's issue types reports. 
    We find the optimal mapping between clusters found by a human annotator and by \Judge, and then define a confusion matrix where each cell $(i, j)$ denotes a number of dataset instances allocated into $i$-th annotator's cluster and $j$-th \Judge's cluster.}
    \label{fig:clusters}
\end{figure*}

\subsection{Examples of per-issue analysis}
Figure~\ref{fig:eval_examples} shows examples of issue explanations generated by GPT-4o and Qwen-2.5-7B.
In the manual inspection of the generated issue explanations, \textit{we observed correct explanations for various kinds of issues}, and rows 1--3 demonstrate such examples. 

We also notice three groups of mistakes. The first occasional problem in per-instance analysis is logical issues. In the example in row 4, the issue is that the ground truth response is not contained as a verbatim substring in the model-generated response, which is a definition of a task metric. However, both GPT-4o and Qwen-2.5-7B claim that the failure in the substring match is caused by the model response containing extra generated information. Such an explanation logically contradicts the task metric, i.e., extra content can only increase chances of finding a given substring in the response, but cannot be a reason for its absence. 

\begin{table}[t!]
  \centering
  \scriptsize
  \begin{tabular}{lllll}
    \toprule
    \textbf{Approach} & \multicolumn{2}{c}{\textbf{Cluster assignment}} & \multicolumn{2}{c}{\textbf{Cluster descriptions}} \\
    \textbf{} & \textbf{ARI$_\text{real}$}  & \textbf{ARI$_\text{syn}$} & \textbf{SLC$_\text{real}$} & \textbf{SLC$_\text{syn}$} \\
    \midrule
    \multicolumn{5}{c}{GPT-4o} \\ \midrule
    Cumulative & 0.14$_{\pm.05}$ & 0.73$_{\pm.05}$ &  0.33$_{\pm.10}$ & 0.70 \\
    Direct & 0.15$_{\pm.05}$ & 0.63$_{\pm.04}$ &  0.42$_{\pm.12}$ & 0.62 \\ \midrule
    \multicolumn{5}{c}{Gemini} \\ \midrule
    Cumulative & 0.13$_{\pm.05}$ & 0.70$_{\pm.02}$ & 0.32$_{\pm.12}$ & 0.71 \\
    Direct & 0.17$_{\pm.04}$ & 0.83$_{\pm.01}$ & 0.42$_{\pm.11}$ & 0.68 \\ \midrule
    \multicolumn{5}{c}{Qwen-2.5-7B} \\ \midrule
    Cumulative & 0.11$_{\pm.04}$ & 0.50$_{\pm.07}$ & 0.41$_{\pm.16}$ & 0.44 \\
    Direct & 0.07$_{\pm.05}$ & 0.01$_{\pm.02}$ & 0.32$_{\pm.11}$ & 0.12 \\ \midrule
    K-means & 0.05$_{\pm.05}$ & 0.44$_{\pm.08}$  & n/a & n/a \\
    Agglomerative & 0.05$_{\pm.00}$ & 0.49$_{\pm.00}$ & n/a & n/a\\
    GMM & 0.04$_{\pm.03}$  & 0.41$_{\pm.05}$  & n/a & n/a \\
    HDBSCAN & 0.01$_{\pm.02}$  & 0.13$_{\pm.03}$ & n/a & n/a \\
    \bottomrule
  \end{tabular}
  \caption{Performance of various approaches and LLMs in issue clustering. Results averaged over 3 runs from different random seeds. Agreement in cluster assignment measured using Adjusted Rank Index (ARI) and in cluster descriptions using LLM-judged Semantic Label Consistency (SLC).
  Subscripts $_{\text{real}}$ and $_{\text{syn}}$ indicate tests on the real and synthetic data respectively. 
 ``N/a'' indicates the metric is not applicable since classic approaches do not generate cluster names.}
  \label{tab:failure-summary-gen}
\end{table}

The second occasional problem in per-instance analysis is the oversimplification of an issue, especially for more unexpected issues, such as an ambiguous task input or an error in evaluation. In the example in row 5, the issue is an ambiguous user question, i.e., both the ground truth and the generated response are correct and provide two different interpretations of the user question. However, GPT-4o and Qwen-2.5-7B report the over-simplified issue of the generated response not providing the same answer as the ground truth response.

Finally, the third occasional reason for a per-instance issue explanation not being accepted by the evaluator LLM, is the subjectivity of some issues in the dataset. For example, a human-annotated issue in row 6, \textit{``The generation was stopped too early because of the reached maximum new tokens limit''}, is evaluated to be not equivalent to the \Judge-generated issue \textit{``The generated response provides an incomplete overview <...>''}. While these two issue explanations are indeed different, they are both correct, and the LLM-generated explanation follows from the annotator's explanation. 

To alleviate potential negative effects from erroneous issue explanations, we recommend developers to check a couple of examples of each issue, which are output by \Judge~in addition to the issue names and descriptions.

\subsection{Issue clustering}

Table~\ref{tab:failure-summary-gen} reports performance in issue clustering, for three LLMs. As described in Section~\ref{sec:evaluation}, we evaluate clustering on both real and synthetic data. 
We find that clustering via direct prompting performs well for small datasets and strong LLMs, but fails for weaker LLM, e.g. Qwen-2.5-7B, and/or larger datasets. For example, ARI reached by GPT-4o on the synthetic data drops from 1 to 0.05 when increasing the dataset size from 100 to 1000 instances. The proposed cumulative clustering demonstrates greater robustness and reaches high ranges of ARI in all cases. In addition, the proposed cumulative algorithm outputs correctly structured summaries by design, while the structural correctness of clustering with direct prompting is not guaranteed. 

Comparing LLMs, we find that Gemini reaches highest scores in both cluster assignment and cluster descriptions generation, followed by GPT-4o and Qwen-2.5-7B. Classic clustering approaches reach rather low values of ARI.

Figure~\ref{fig:clusters} demonstrates examples of confusion matrices for several datasets. Pronounced diagonals and matching cluster names illustrate the strong capabilities of \Judge~to output issue types reports that resemble the issue reports produced by humans. Due to the inherent subjectivity of clustering task, we observe occasional merging or splitting of annotator's clusters, e.g. clusters \textit{``Wrong topping''} and \textit{``Wrong variables''} were merged by \Judge~into one cluster \textit{``Entity Mislabeling''} for the Pizza ordering dataset. 

\textit{To sum up, our results demonstrate the effectiveness of the proposed cumulative clustering approach to produce issue reports that resemble the ones produced by humans. For practical applications, we recommend using recent models such as GPT-4o or Gemini-2.5-Flash.}

\section{Discussion}
\label{sec:discussion}
In this section, we discuss potential extensions of the proposed approach.
\paragraph{Issues prefiltering.} As discussed in Section~\ref{sec:method}, \Judge~operates only over instances which received low scores from a quantitative task metric. Such prefiltering could in principle be 
removed and incorporated directly into per-instance analysis by modifying its prompt, e.g. ``\textit{Output what is an issue with this instance. If there is no issue, output 'No issue'}''. Instances with predictions ``\textit{No issue}'' then would be discarded from issue clustering. However, in preliminary experiments we found that LLM are prone to making up issues for fully correct instances. Hence, we do not recommend removing the prefiltering step (at least with the current state of LLMs), which is a reasonable design since \Judge~is an \textit{error analysis} method.

\paragraph{Multiple issues per instance.} \Judge~could be easily extended to detect multiple issues per instance, by modifying the prompt used for per-instance analysis and going through the generated issues one-by-one in the issue clustering step. However, same as with a previous discussion point, in our preliminary experiments we found that LLMs are prone to generating non-existing issues in such a scenario. For example, \verb|GPT-4o| tends to generate a constant number of issues for any instance (in particular, 3). 
In practice, we believe that our design with one issue per instance is reasonable, since most of the erroneous instances have only one issue. Furthermore, even if some instances have repeating issues, our algorithm would still capture most of the issues in the dataset due to issue repetition.

\paragraph{Pairwise comparison of models.} \Judge~can be straightforwardly extended to perform pairwise comparison of models, by running per-instance analysis for outputs of both models and using two counters (one per model) in the issue clustering step.

\paragraph{Use without ground truth labels.} \Judge~can be straightforwardly used without ground truth labels, i.e. as an unsupervised evaluation metric, if the LLM internal knowledge is sufficient to understand errors in a given task. \Judge~can also be provided with additional evaluation metadata, e.g. score rubrics used in quantitative evaluation.

\section{Related work}
\paragraph{Quantitative LLM-based evaluation.} 
While using commercial LLMs for evaluation remains common practice, one line of work~\citep{prometheus, prometheus2, multiprometheus} focuses on tuning open-source LLMs on the synthetically generated evaluation data, to ensure reproducibility of evaluation. Other works improve quantitative LLM-as-a-judge by conducting more fine-grained evaluation, e.g. using multiple evaluation criteria~\citep{liang2023holistic,fu-etal-2024-gptscore} or selecting evaluation criteria individually per instance~\citep{flask,checklists,biggen}. 
Composite evaluation approaches such as FactScore~\citep{factscore} or RAGChecker~\citep{ragchecker} use LLMs in the intermediate evaluation steps.

\paragraph{Qualitative LLM-based evaluation.} LLM-generated \textit{qualitative} error explanations are often used to improve the precision of quantitative evaluation~\citep{zeng2024evaluating,flask} or to explain the assigned quantitative scores to a developer~\citep{instructscore,tigerscore}. Such approaches only output \textit{per-instance} explanations, and a \textit{substantial human effort is still needed to read all of them}. Certain works~\citep{tigerscore,factgenie,matese} focus on outputting aggregated reports of frequent errors, but with a (limited) predefined error set, i.e. they solve the task of error classification. In contrast to these efforts, \Judge~outputs an \textit{aggregated} report of issue types discovered in an \textit{open-ended} manner, i.e. without any predefined issue set.

\paragraph{Meta-evaluation.} A line of community efforts~\citep{zeng2024evaluating,lambert-etal-2025-rewardbench,metal,bavaresco2024llmsinsteadhumanjudges} is devoted to an important task of meta-evaluating LLM-as-a-judge, i.e. collecting human annotations for various tasks, domains, or languages, and evaluating how closely LLMs mirror human judgments. Certain task-specific datasets~\citep{freitag-etal-2021-experts} can be used to meta-evaluate fine-grained issue detection. Our work further contributes to this direction by the release of a meta-evaluation dataset, containing \textit{qualitative} issue explanations for 12 datasets from 7 tasks and their per-dataset \textit{clustering}.

\paragraph{Clustering with LLMs.}
Earlier works~\citep{DBLP:journals/corr/abs-2403-15112,DBLP:journals/corr/abs-2405-07278} demonstrate advantages of leveraging LLM-derived embeddings in place of traditional TF-IDF or BERT vectors in standard clustering algorithms. 
More recent works employ LLMs directly to cluster textual data. \citet{DBLP:journals/tacl/0002GGLWN24} instruct a GPT-3.5 model to cluster the provided data given few-shot demonstrations. \citet{DBLP:journals/corr/abs-2410-00927} transform clustering into a two-stage classification task: first prompting an LLM to infer a set of candidate clusters for the dataset, then prompting it to assign the best cluster to each instance. ClusterLLM~\cite{DBLP:conf/emnlp/0001WS23} uses an instruction-tuned LLM to guide clustering, i.e., to decide which clusters to merge. In our work, we propose an alternative intuitive approach for LLM-based clustering. Our approach can also be extended in the future with the listed strategies.

\section{Conclusion}
In this work, we present \Judge, a novel approach for generating structured reports summarizing key types of issues in a given NLG system. We hope that this approach will help developers to spot more easily issues and artifacts in their NLG systems. 

Future works could equip \Judge~with advanced reasoning or agentic pipelines, tune LLMs for issue report generation, and study the approach for a wider set of languages.

\section*{Limitations}
As any LLM-based system, \Judge~can make occasional mistakes in analysis or clustering. In Section~\ref{sec:results}, we discuss types of such mistakes and recommend checking a couple of examples of each issue, which are also output by \Judge.  

Regarding limitations of the evaluation methodology, despite our efforts in considering a diverse set of tasks, domains, and LLMs, we acknowledge the infeasibility of covering the entire breadth of NLG applications and models in our study. Another limitation is that we mainly focus on English. We believe our findings will transfer to other languages, with the use of strong recent multilingual LLMs, but acknowledge that the reliability of \Judge~in multilingual studies requires a separate study.

\section*{Broader impact}
We acknowledge that as any LLM-based system, \Judge~can make errors which could propagate to the downstream systems and decrease their performance. For example, if developers rely solely on the issue names formulated by Judge, this could occasionally lead to unnecessary or even harmful modifications of their NLG systems. This could also happen in case of misinterpetation of an issue by a developer due to issue subjectivity. To reduce such risks, we recommend developers to check examples of issue types, which are also output by \Judge, in addition to the issue names and description. 

\section*{Acknowledgments}
We greatly appreciate the help of Alexandre Misrahi, Salah Aït-Mokhtar, and Maxime Louis. 
The project was initiated at the Advanced Language Processing School (ALPS 2025, \url{https://alps.imag.fr}).

A part of this work was carried out within the framework of the AugmentIA Chair, supported by the Fondation Grenoble INP thanks to the patronage of Artelia Group, and is affiliated with Laboratory of informatics in Grenoble (LIG).
A part of this work received government funding managed by the French National Research Agency under France 2030, reference ANR-23-IACL-0006.

\nocite{misrahi2025adaptinglargelanguagemodels}
\bibliography{custom}

\appendix
\newpage
\onecolumn

\section{Details on data annotation}
\label{sec:annotation}

\paragraph{Per-instance analysis.} 
The core of our meta-evaluation strategy is to collect manual annotations of failure cases for a set of instances from various tasks and domains. 
For each instance, consisting of a task input, a ground truth response, a generated response, a description of a task metric, and optionally retrieved documents, an annotator’s task is to formulate what the particular issue is in this instance. We then prompt an \textit{evaluator LLM} to judge whether the issue explanation determined by the \Judge~for a particular instance matches the issue determined by the human annotator. The outputs from the \textit{evaluator LLM} are binary and can be accumulated into a \textit{per-instance analysis accuracy} score. 

The annotation instruction asks to ignore instances which have multiple issues (to avoid ambiguity in per-instance analysis), instances where ground truth labels appear to be wrong, and instances where the annotator’s expertise is insufficient to judge the correctness of the generated answer. We also limit the number of instances with the same issue to not exceed 8 examples per dataset, to ensure the diversity of the final dataset. 

\paragraph{Issue clustering.} Human annotation also includes a step of manually clustering issues discovered in per-instance analysis, i.e., specifying cluster indices and cluster names (generalized issue types) for labeled instances. This annotation is then used to compute clustering agreement between the clustering produced by \Judge~and by a human annotator.

\paragraph{Dataset composition.}
For each considered dataset, we manually label failures in up to 40 generations from one of the open-source LLMs (\verb|Qwen-2.5-1b|, \verb|Llama-3.2-1b|, \verb|Command-R-35b|, \verb|Vicuna-1.5-13B|). 

\paragraph{Annotation details.}
All the labeling was performed by the authors of the paper in Google Spreadsheets\footnote{\url{https://docs.google.com/spreadsheets}}. Each instance was annotated by one author. Authors of the paper are PhD students in the NLP field or have already completed their PhD in NLP and are employed as NLP researchers. 

Time needed for data annotation varies between tasks: it took us 1—6 hours per task. Tasks with longer inputs, e.g. RA-QA, and from more complex domains, e.g. biomedical, take more time to annotate, e.g. they require reading the retrieved documents carefully.

Below is an annotation instruction:
\begin{figure*}[ht]
\begin{tcolorbox}[colback=gray!2, colframe=blue!60, width=\textwidth]
\texttt{For each example, consisting of a user prompt, a ground truth label, an LLM generation, and optionally retrieved documents, an annotator’s task is to formulate what is a particular failure case in this example. Identify only one, most important specific, clearly visible issue in each test case. Please formulate the detected issue as a clear, full sentence, e.g. "The generated response is in German instead of French which is the language of the user input" or "The retrieved documents are from a datastore which is irrelevant to the given user question".}

\texttt{Please add your annotations in the following Google Spreadsheet: [link], column “Per-instance analysis”. You can skip instances (rows) for which you feel that you do not have enough expertise to detect an issue, which have multiple issues, or for which ground truth labels appear to be wrong.}

\texttt{After annotating per-instance analysis, suggest a clustering of the detected issues, i.e. how would you group them, and add the corresponding cluster indexes and names in columns “Cluster index” and ``Cluster name''.}
\end{tcolorbox}
  \label{fig:annotation_instruction}
\end{figure*}

\paragraph{Inter-annotator agreement.} We measure the inter-annotator agreement on a subset of 100 instances, i.e. each of these instances was labeled by two annotators and then we computed their agreement using the same evaluator LLM as in other experiments, i.e. Claude-3.7-sonnet. The resulting inter-annotator agreement was 57\% (percentage of cases when two annotators suggested the same issue, as judged by Claude-3.7-sonnet), i.e. the similar range as the scores we obtain in Table~\ref{tab:per_example_analysis}. 

The main factor contributing to the moderate agreement is the subjectivity of issue analysis. For example, in a situation when generation was stopped due to reaching the maximum new tokens limit, one annotator said “The response is incomplete” and another annotator said “Generation was stopped too early”. Both denote the same root issue, but are formulated differently and Claude judges these comments as different.

\section{Further details on the experimental setup}
\label{sec:expsetup}
\paragraph{Case study.} For each of the three considered BigBenchHars tasks, we build a simple initial generative pipeline. This pipeline is then improved in two rounds by generating issue reports with \Judge. Configurations of the initial pipeline are as follows. System prompt: ``\textit{You are a helpful assistant. Output your answer after a final separator `Answer:'}''.
Generation hyperparameters: all hyperparameters set to default values from the HuggingFace or OpenAI API, plus setting maximum new tokens or 500 for HuggingFace models. The final answers are obtained by cropping the content after a final separator ``\textit{Answer:}'' and applying a \verb|.strip()| python function. Evaluation function: exact match with ground truth. \Judge~is run with GPT-4.1 and providing a one-sentence description of a task metric, i.e. ``Evaluation is conducted using exact matching between the ground-truth label and the content of the generated response after the final separator `Answer:'''.

\paragraph{Meta-evaluation experiments.} 
For each instance, \Judge~is provided with a task input, a ground truth response, a generated response, 5 retrieved documents (only for RA-QA), and a short task comment. The task comment describes the task metric (in one sentence), provides a comment on the nature of ground truth responses (either that it is the expected answer or that it is only one of the possible correct answers), and also contains a comment that retrieval-augmented generation (RAG) or Chain-of-Thought (COT) prompting was used, when applicable (6 datasets with RAG and 2 datasets with COT). The used task metrics are binary LLM-as-a-judge (the generated response is accepted or not) or binary Match (outputs \verb|True| is one of the ground truth answers is contained a substring in the generated response, and \verb|False| otherwise). Task comments for all datasets are also presented in Appendix~\ref{app:examples}. 

For \Judge, open-source LLMs are run on a single V100 GPU with greedy decoding ($\sim$20 GPU-hours in total). Commercial LLMs are run via API with requesting \verb|json| output format.

The time of running the \Judge~algorithm depends on the setting (commercial vs open-source LLMs, type of GPU etc) and in our experiments was taking 2—30 min, i.e. reasonably short on the scale of the time needed to develop an NLG system.

\paragraph{Meta-evaluation of an evaluator LLM.} To ensure the reliability of the \textit{evaluator LLM}, we collected a small meta-evaluation dataset of 50 instances from 4 datasets (MKQA (ru), RobustQA Writing, FLASK, MultiDetox), where the equivalence of the \Judge’s and \textit{human annotator’s} per-instance analysis was judged by a human annotator and can be compared to the \textit{evaluator LLM}’s verdict. Strong commercial LLMs, such as \verb|GPT-4o|, \verb|Gemini-2.0-Flash|, and \verb|Claude-3.7-Sonnet|, achieved a meta-evaluation accuracy of 85-90\% on this dataset, and an open-source \verb|Solar-10.7B|~\citep{kim2023solar} achieved a meta-accuracy of 60\%. In all the experiments, we use \verb|claude-3-7-sonnet-20250219| as the \textit{evaluator LLM}, to avoid using the same LLM for analysis and for evaluation.

\section{Clustering experiment setup}
\label{sec:clustering-setup}

In this experiment, we perform a hyperparameter grid search for five clustering algorithms: KMeans, Agglomerative Clustering, Spectral Clustering, Gaussian Mixture Models (GMM), and HDBSCAN on a synthetic set. Each algorithm is evaluated across a range of hyperparameter combinations. 
For KMeans, we vary the \verb|distance_metric| (euclidean, cosine), \verb|kmeans_init| strategy (k-means++, random), \verb|kmeans_n_init| (10, 50), and \verb|kmeans_max_iter| (300, 500). For Agglomerative Clustering, we test all combinations of \verb|distance_metric| (euclidean, cosine) and \verb|linkage_type| (ward, average, complete), while ensuring that ward is only paired with euclidean (as required by the algorithm).  Spectral Clustering configurations include \verb|distance_metric| (euclidean, cosine), \verb|assign_labels| (kmeans, discretize), \verb|spectral_gamma| (0.1, 0.5, 1.0, 2.0), and \verb|spectral_n_neighbors| (5, 10, 20). 
For GMM, we explore \verb|covariance_type| (full, diag), \verb|gmm_init_params| (kmeans, random), and \verb|gmm_max_iter| (100, 300). 
Lastly, HDBSCAN is tested with \verb|distance_metric| (euclidean, cosine), \verb|min_cluster_size| (3, 5, 10, 15, 20), \verb|hdbscan_min_samples| (None, 1, 5), and \verb|hdbscan_cluster_selection_method| (eom, leaf). 
Each valid configuration is evaluated over three independent trials with different random seeds to ensure robustness. After collecting results based on Adjusted Rand Index (ARI), the best-performing configuration for each algorithm on the synthetic validation set is selected. These best configurations are then applied to the test set of synthetic data and to the real dataset.
\clearpage

\section{Models}
\label{sec:models}

\begin{table}[h]
\centering
\scriptsize
\begin{tabular}{p{2cm}p{2.5cm}p{3.5cm}p{6cm}}
    \toprule
    \textbf{Model} & \textbf{BibTeX} & \textbf{License} & \textbf{Model Repository} \\
    \midrule
    GPT-4o & 
    \cite{openai2024gpt4technicalreport} & 
    Proprietary & 
    \url{https://platform.openai.com/docs/models/gpt-4o} \\
    Gemini-2.0-Flash & 
    \cite{geminiteam2025geminifamilyhighlycapable} & 
    Proprietary & 
    \url{https://deepmind.google/technologies/gemini/flash/}\\
    Qwen-2.5 & 
    \cite{qwen2025qwen25technicalreport} &
    Apache 2.0 & 
    \url{https://huggingface.co/collections/Qwen/qwen25-66e81a666513e518adb90d9e} \\
    DeepSeek-R1-Distill-Llama-8B & \cite{deepseekai2025deepseekr1incentivizingreasoningcapability} & 
    Llama 3.1 Community License & 
    \href{https://huggingface.co/deepseek-ai/DeepSeek-R1-Distill-Llama-8B}{https://huggingface.co/deepseek-ai/DeepSeek-R1-Distill-Llama-8B} \\
    Aya-Expanse-8B & 
    \cite{dang2024ayaexpansecombiningresearch} & 
    Creative Commons Attribution Non Commercial 4.0 & 
    \href{https://huggingface.co/CohereLabs/aya-expanse-8b}{https://huggingface.co/CohereLabs/aya-expanse-8b} \\
    Llama-3.1-8B-Instruct & 
    \cite{grattafiori2024llama3herdmodels} & 
    Llama 3.1 Community License & 
    \href{https://huggingface.co/meta-llama/Llama-3.1-8B-Instruct}{https://huggingface.co/meta-llama/Llama-3.1-8B-Instruct} \\
    Llama-3.2-1B-Instruct & 
    \cite{grattafiori2024llama3herdmodels} & 
    Llama 3.2 Community License & 
    \href{https://huggingface.co/meta-llama/Llama-3.2-1B-Instruct}{https://huggingface.co/meta-llama/Llama-3.2-1B-Instruct} \\
    Ministral-8B-Instruct & 
    \cite{MistralAI2024Ministraux} & 
    Mistral AI Research License & 
    \href{https://huggingface.co/mistralai/Ministral-8B-Instruct-2410}{https://huggingface.co/mistralai/Ministral-8B-Instruct-2410} \\
    Solar-10.7B & 
    \cite{kim2023solar} & 
    Creative Commons Attribution Non Commercial 4.0 & 
    \href{https://huggingface.co/upstage/SOLAR-10.7B-Instruct-v1.0}{https://huggingface.co/upstage/SOLAR-10.7B-Instruct-v1.0} \\
    Vicuna-1.5-13B & \cite{vicuna2023} & Llama 2 Community License Agreement & \url{https://huggingface.co/lmsys/vicuna-13b-v1.5} \\
    Command-R-35B & \cite{commandr} & Creative Commons Attribution Non Commercial 4.0 & \url{https://huggingface.co/CohereLabs/c4ai-command-r-v01} \\
    \bottomrule
\end{tabular}
\caption{References to the used LLMs; all LLMs allow use for research.}
\label{tab:llm_info}
\end{table}

\section{Datasets}
\label{sec:datasets-expe}

\begin{table}[h!]
  \centering
  \scriptsize
  \begin{tabular}{p{4.cm}p{3.5cm}p{6.5cm}}
    \toprule
    \textbf{Dataset name} & \textbf{Dataset reference} & \textbf{License} \\
    \midrule
    \multicolumn{3}{l}{\textbf{Natural Language Generation}} \\
    \midrule
    FLASK data mix: Self-Instruct, WizardLM, Koala, CommonSense QA & \cite{wang-etal-2023-self-instruct, xu2024wizardlm, koala_blogpost_2023, DBLP:conf/nips/TalmorYBBGCB21} & Apache 2.0, MIT, Apache 2.0, Creative Commons Attribution 4.0 \\
    WMT'22 & \cite{kocmi-etal-2022-findings} & Apache 2.0 \\
    Elitr-Bench & \cite{DBLP:conf/coling/ThonetBR25} & Attribution 4.0 International \\
    PIZZA & \cite{DBLP:journals/corr/abs-2212-00265} & Attribution-NonCommercial 4.0 International \\
    GSM8K & \cite{DBLP:journals/corr/abs-2110-14168} & MIT License \\
    ParaDetox & \cite{dementieva2024overview} & OpenRAIL++ \\
    \midrule
    \multicolumn{3}{l}{\textbf{Retrieval-augmented QA}} \\
    \midrule
    MKQA (ru) & \cite{longpre-etal-2021-mkqa} & Creative Commons Attribution-ShareAlike 3.0 Unported License \\
    BioASQ & \cite{krithara2023bioasq} & Attribution 2.5 Generic \\
    RobustQA & \cite{lotte,DBLP:conf/emnlp/Han00XWLWMC24,Han2023} & Apache-2.0 \\
    SearchQA & \cite{DBLP:journals/corr/DunnSHGCC17} & BSD 3-Clause  \\
    SyllabusQA & \cite{DBLP:conf/acl/FernandezSL24} & Attribution-NonCommercialShareAlike \\
    \midrule
    \multicolumn{3}{l}{\textbf{BigBenchHard}} \\
    \midrule
    Date understanding; Word sorting; Movie recommendation & \cite{bbh} & MIT \\
    \bottomrule
  \end{tabular}
  \caption{References to the used datasets; all datasets allow use for research. We select instances from test splits.}
  \label{tab:dataset_licenses}
\end{table}

\section{Per-dataset results}
Table~\ref{tab:per_dataset_analysis} presents per-dataset results for for GPT-4o as \textit{LLM-as-a-qualitative-judge}. 

\begin{table}[t!]
  \centering
  \scriptsize
  \begin{tabular}{p{2.3cm}>{\raggedleft\arraybackslash}p{1.3cm}>{\raggedleft\arraybackslash}p{1.1cm}>{\raggedleft\arraybackslash}p{1.1cm}}
    \toprule
    \textbf{Dataset} & \textbf{Per-inst. an. acc. (\%)} & \textbf{Issue clust. ARI} &  \textbf{Issue clust. SLC}\\
    \midrule
    Semantic parsing & 94.1 & 0.41 & 0.29 \\ 
    Grade school math & 88.2 & 0.04 & 0.22 \\ 
    Detoxification & 77.8 & 0.36 & 0.28 \\ 
    Long-context QA & 69.2 & 0.07 & 0.50 \\ 
    Translation en-ru & 65.8 & 0.10 & 0.63 \\ 
    Instruction following & 55.9 & 0.09 & 0.19  \\ 
    RA-QA: SyllabusQA & 77.8 & 0.16 & 0.55 \\ 
    RA-QA: MKQA (ru) & 75.8 & 0.17 & 0.44 \\ 
    RA-QA: BioASQ & 66.7 & 0.08 & 0.31 \\ 
    RA-QA: SearchQA & 38.4 & 0.15 & 0.16 \\ 
    RA-QA: Writing & 30.7 & 0.00 & 0.11 \\ 
    RA-QA: Lifestyle & 19.0 & 0.00 & 0.32 \\
    \bottomrule
  \end{tabular}
  \caption{Per-dataset results for GPT-4o as \textit{LLM-as-a-qualitative-judge}.}
  \label{tab:per_dataset_analysis}
\end{table}

\section{Prompts}
\label{sec:prompts}

\begin{figure*}[ht]
\begin{tcolorbox}[colback=gray!2, colframe=blue!60, width=\textwidth]
\texttt{You are an expert in analysing the failure cases in natural language generation tasks \\
You are given a question, ground truth label(s), and the answer generated by an LLM. \\
The generated answer was not accepted by the automatic evaluation measured with metric \{metric info\} \\
Read all these materials and reply what is the particular failure case in this example, i.\,e. why exactly the generated response was not accepted. \\
The problem can be in any part of the pipeline, including the question itself or any aspects of the system outputs. \\
IMPORTANT: identify ONE, MOST IMPORTANT, SPECIFIC, CLEARLY VISIBLE issue in each test case. \\}

\vspace{0.5em}

\texttt{Question: \\
*** \\
\{question\} \\
*** \\}

\vspace{0.5em}

\texttt{Ground truth label(s): \\
*** \\
\{label\} \\
*** \\ }

\vspace{0.5em}

\texttt{LLM-generated answer: \\
***\\
\{answer\}\\
***\\}

\vspace{0.5em}

\texttt{So what is the particular failure in this example? First output a detailed analysis, and then output a final summary of the failure in one or two sentences after a special separator ``Summary:''.}
\end{tcolorbox}
 \caption{Prompt used for per-instance analysis. The presented version of the prompt is for text LLM outputs, the prompt can be easily changed if JSON outputs are supported by an LLM. For RA-QA, we also include retrieved documents in the prompt.}
  \label{fig:per_ex_analysis_prompt}
\end{figure*}

\begin{figure*}[ht]
\begin{tcolorbox}[colback=gray!2, colframe=blue!60, width=\textwidth]
\texttt{You are an expert in analysing the failure cases of natural language generation systems.  \\
You already performed per-example analysis, where for each example, you were given a question, ground truth label(s), and the answer generated by an LLM.\\
The generated answer in all examples was not accepted by the automatic evaluation. \\
You already read all these materials and formulated what was the particular failure case in each example, i.e. which part of the pipeline failed so that the generated response was not accepted. \\}

\vspace{0.5em}

\texttt{Now your task is to summarize all your per-example analyses into a concise overall summary of failure cases for the given dataset. \\ 
Summarize what are the various failure types in this dataset (provide the overall count of each error type and also ids of all examples of each error type). \\
Please try to be very specific in determining error types, avoid too much generic error types. On the contrary, determine as much as possible FINE GRAINED error types.  \\
Furthermore, provide a comment for each error type explaining the essence of this error type in a bit more details (in the context of this particular dataset). \\}

\vspace{0.5em}

\texttt{*** Per-example analysis which you generated before ***\\
\{analysis\}\\
*** per-example analysis ended ***\\}

\vspace{0.5em}

\texttt{Summarize all your per-example analyses into a concise overall summary of failure cases.\\
Generate a json with the only key ``summary'', and a value is a dict of error types. Each value in this dict (corresponding to one detected error type) is a dictionary with keys ``error\_name'', ``error\_description'', ``indexes'' (indexes of all examples with this error type), and ``num\_examples'' (overall count of this error type).}
\end{tcolorbox}
 \caption{Prompt used for summary direct prompting.}
  \label{fig:summary_prompt}
\end{figure*}

\begin{figure*}[ht]
\begin{tcolorbox}[colback=gray!2, colframe=blue!60, width=\textwidth]
\texttt{You are an expert in analysing the failure cases of natural language generation systems.  \\
You already performed per-example analysis, where for each example, you were given a question, ground truth label(s), and the answer generated by an LLM. \\
The generated answer in all examples was not accepted by the automatic evaluation. \\
You already read all these materials and formulated what was the particular failure case in each example, i.\,e. which part of the pipeline failed so that the generated response was not accepted.\\}

\vspace{0.5em}

\texttt{Now your task is to summarize all your per-example analyses into a concise overall summary of failure cases for the given dataset.  \\
You are using a SPECIAL CUMULATIVE ALGORITHM as follows. You are going through examples one by one and accumulate discovered error cases in a special pool. For each example, you check if any of the already discovered error types from the pool fits it, and if so, you assign this error type to this example. If none of already existing error types fit the current example, you create a new error type and add it to a pool. \\}

\vspace{0.5em}

\texttt{A pool of already discovered error cases: \\
***\\
\{error cases\}\\
***\\}

\vspace{0.5em}

\texttt{Analysis of a current example: \\
*** \\
\{analysis\} \\
*** \\}

\vspace{0.5em}

\texttt{Do you think any of the already discovered error cases fit the current example? If yes, output ``Decision:'' and a key marking the chosen error case, e.\,g. ``Decision: type\_0''. Do it only if you are REALLY sure that the chosen error case fits the current example! DO NOT output cluster name. If not, output ``Decision: None''. DO NOT output anything else.}
\end{tcolorbox}
 \caption{Prompt used for classifying each instance in the cumulative clustering strategy. The presented version of the prompt is for text LLM outputs, the prompt can be easily changed if JSON outputs are supported by an LLM.}
  \label{fig:summary_cum_strategy_prompt}
\end{figure*}

\begin{figure*}[ht]
\begin{tcolorbox}[colback=gray!2, colframe=blue!60, width=\textwidth]

\texttt{You are an expert in analysing the failure cases of natural language generation systems. \\
You already performed per-example analysis, where for each example, you were given a question, ground truth label(s), and the answer generated by an LLM. \\
The generated answer in all examples was not accepted by the automatic evaluation. \\
You already read all these materials and formulated what was the particular failure case in each example, i.\,e. which part of the pipeline failed so that the generated response was not accepted. \\
Now your task is to summarize all your per-example analyses into a concise overall summary of failure cases for the given dataset. \\ 
You are using a SPECIAL CUMULATIVE ALGORITHM as follows. You are going through examples one by one and accumulate discovered error cases in a special pool. For each example, you check if any of the already discovered error types from the pool fits it, and if so, you assign this error type to this example. If none of already existing error types fit the current example, you create a new error type and add it to a pool. \\}

\vspace{0.5em}

\texttt{A pool of already discovered error cases: \\
*** \\
\{error cases\} \\
*** \\}

\vspace{0.5em}

\texttt{Analysis of a current example: \\
*** \\
\{analysis\} \\
***\\}

\vspace{0.5em}

\texttt{You decided to create a new error type for a given example, not yet present in a pool. Now you need to generate SHORT LABEL and a 1 or 2 SENTENCE DESCRIPTION for this new error type. Please try to be very specific in determining a FINE GRAINED error type, avoid too much generic error types. At the same time, it is important that the generated label and description of the error type can be GENERALIZED to other examples, i.\,e. avoid references to the particular content of the current example (names, dates, etc): anything related ONLY to this example SHOULD NOT be present in the description and label.\\
Output answer is the following format: "{SHORT LABEL}: {DESCRIPTION}", do not output anything else!}
\end{tcolorbox}
 \caption{Prompt used for generating a new issue type in the cumulative clustering strategy. The presented version of the prompt is for text LLM outputs, the prompt can be easily changed if JSON outputs are supported by an LLM.}
  \label{fig:summary_rag_cumulative_newtype_prompt}
\end{figure*}

\newpage
\clearpage
\begin{figure*}[ht]
\begin{tcolorbox}[colback=gray!2, colframe=blue!60, width=\textwidth]
\texttt{Situation: Two experts are inspecting examples in natural language generation for a particular dataset.  \\
You will be given 2 sentences, which represent the conclusions of the two experts about the same example, i.\,e. what is a failure case in the example they were given.\\
Your task is to determine if the experts describe the same failure. \\}

\vspace{0.5em}

\texttt{Expert 1 conclusion: E1 \\ 
Expert 2 conclusion: E2\\}

\vspace{0.5em}

\texttt{Do the experts describe the same failure? Output only one word `Yes' or `No'.}
\end{tcolorbox}
 \caption{Prompt used for evaluation. The presented version of the prompt is for text LLM outputs, the prompt can be easily changed if JSON outputs are supported by an LLM.}
  \label{fig:evaluation_prompt}
 \end{figure*}

\section{Examples}
\label{app:examples}
The following pages present examples of task instances, per-instance analysis, generated error reports, and clustering confusion matrices, for all 12 considered datasets. Clusters of size 1 are shown in confusion matrices but omitted in error reports, for space purposes.

\newpage
\pagenumbering{gobble}

\subsection{Semantic parsing (PIZZA dataset)}
\center{\includegraphics[height=23cm]{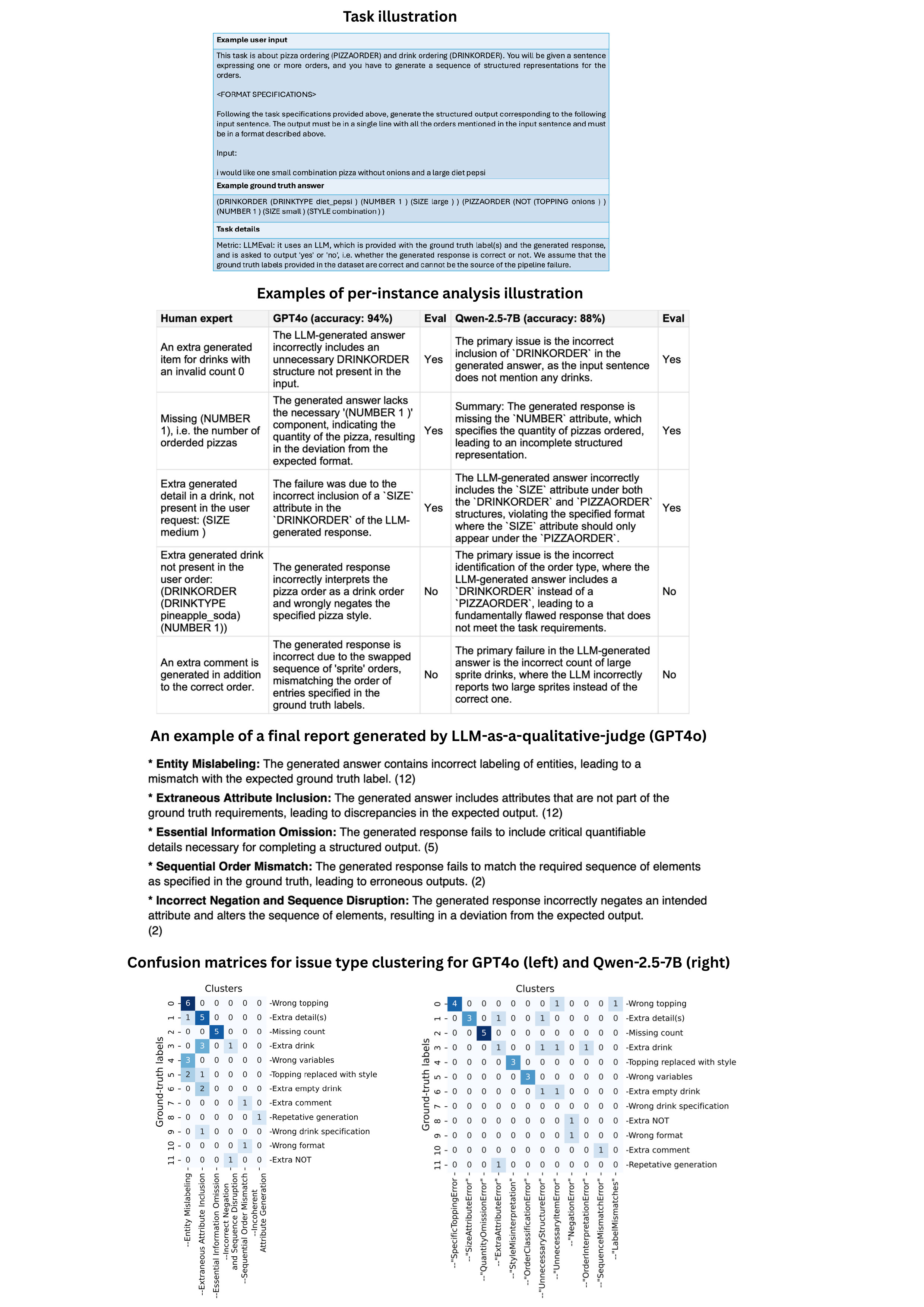}}

\subsection{Long context QA (Elitr-Bench dataset)}
\center{\includegraphics[height=24cm]{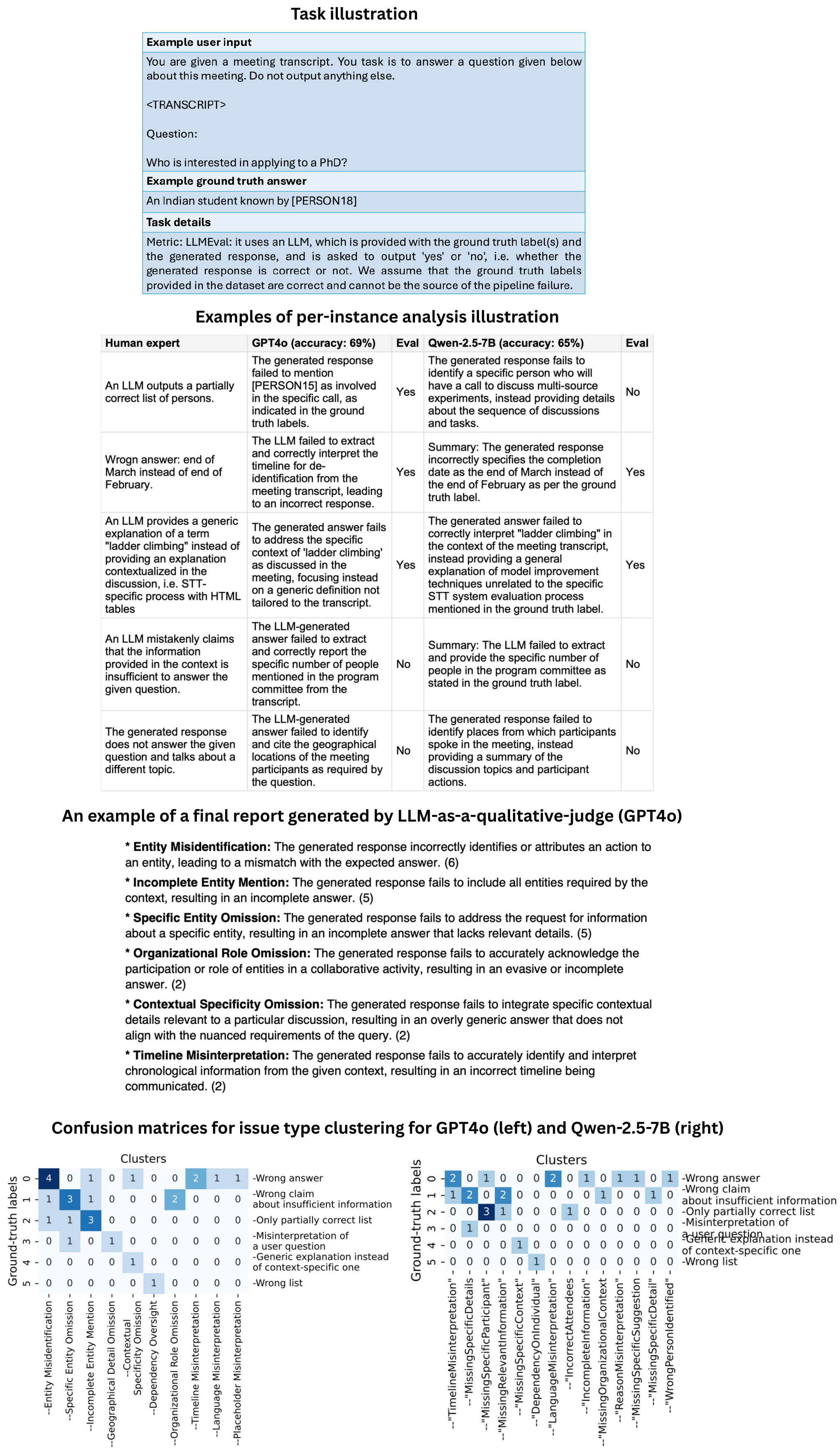}}

\subsection{Detoxification (MultiDetox dataset)}
\center{\includegraphics[height=24cm]{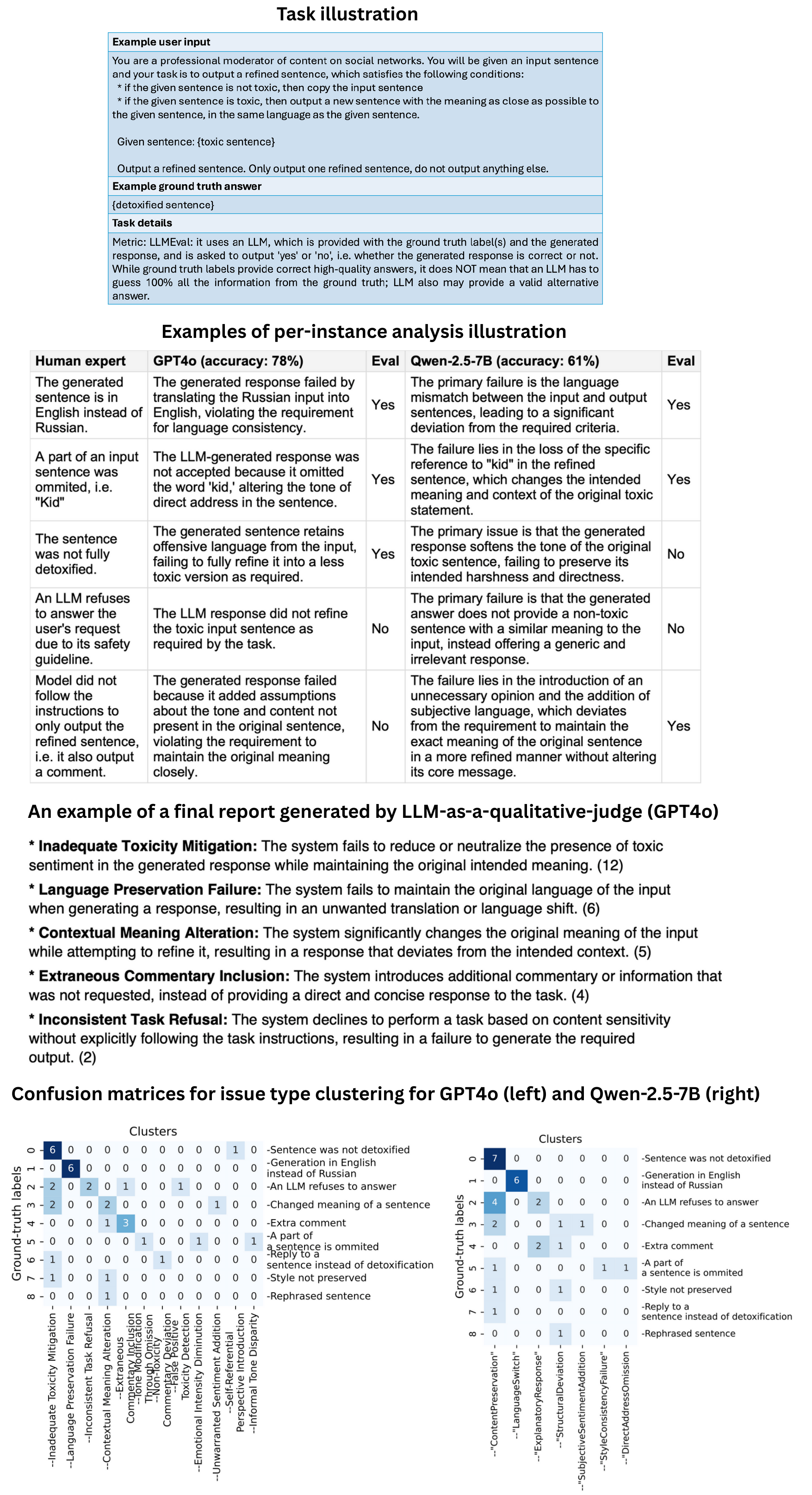}}

\subsection{Translation en-ru (WMT'22 dataset)}
\center{\includegraphics[height=24cm]{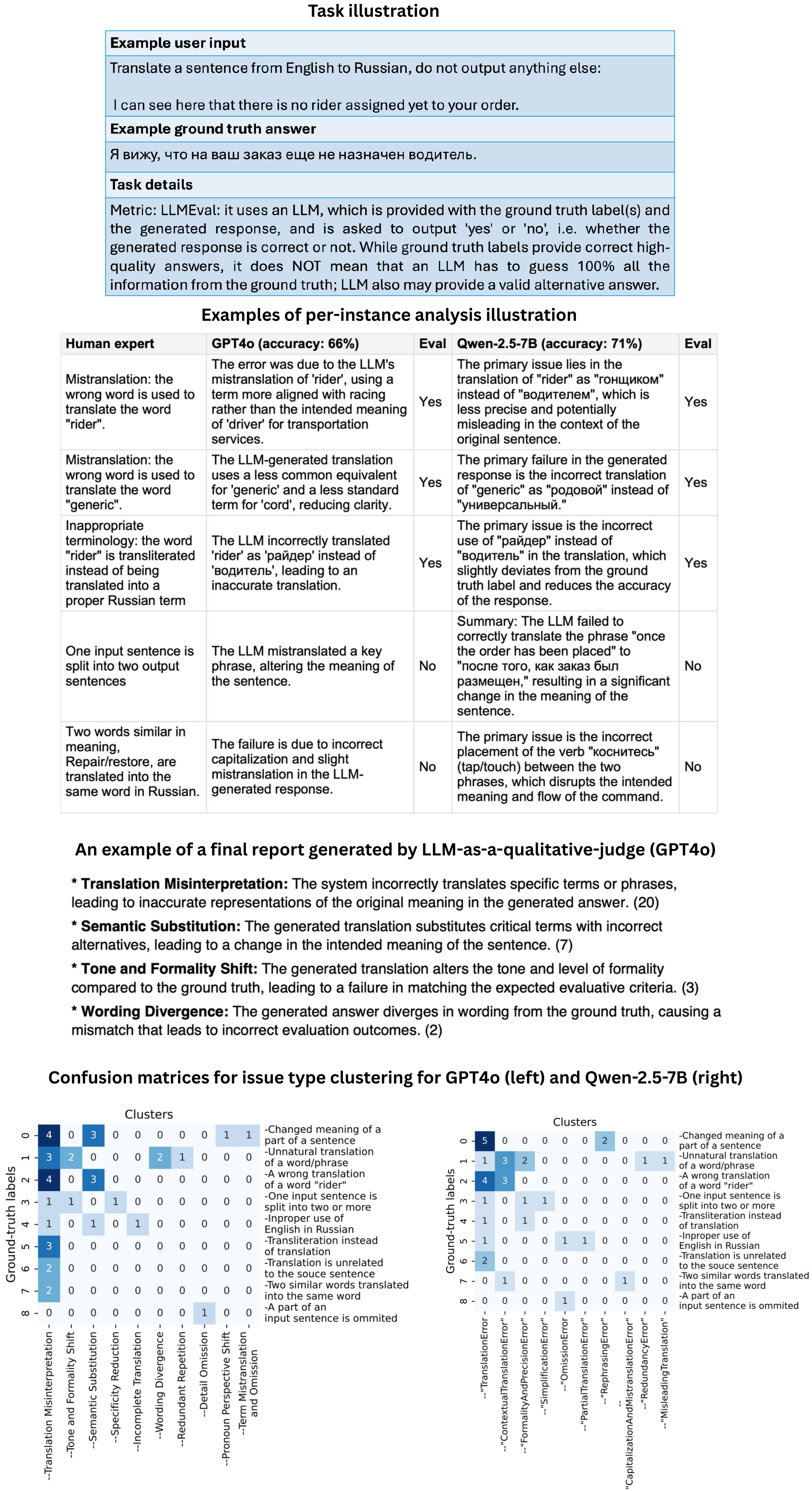}}

\subsection{Instruction following (FLASK dataset)}
\center{
    \includegraphics[height=24cm]{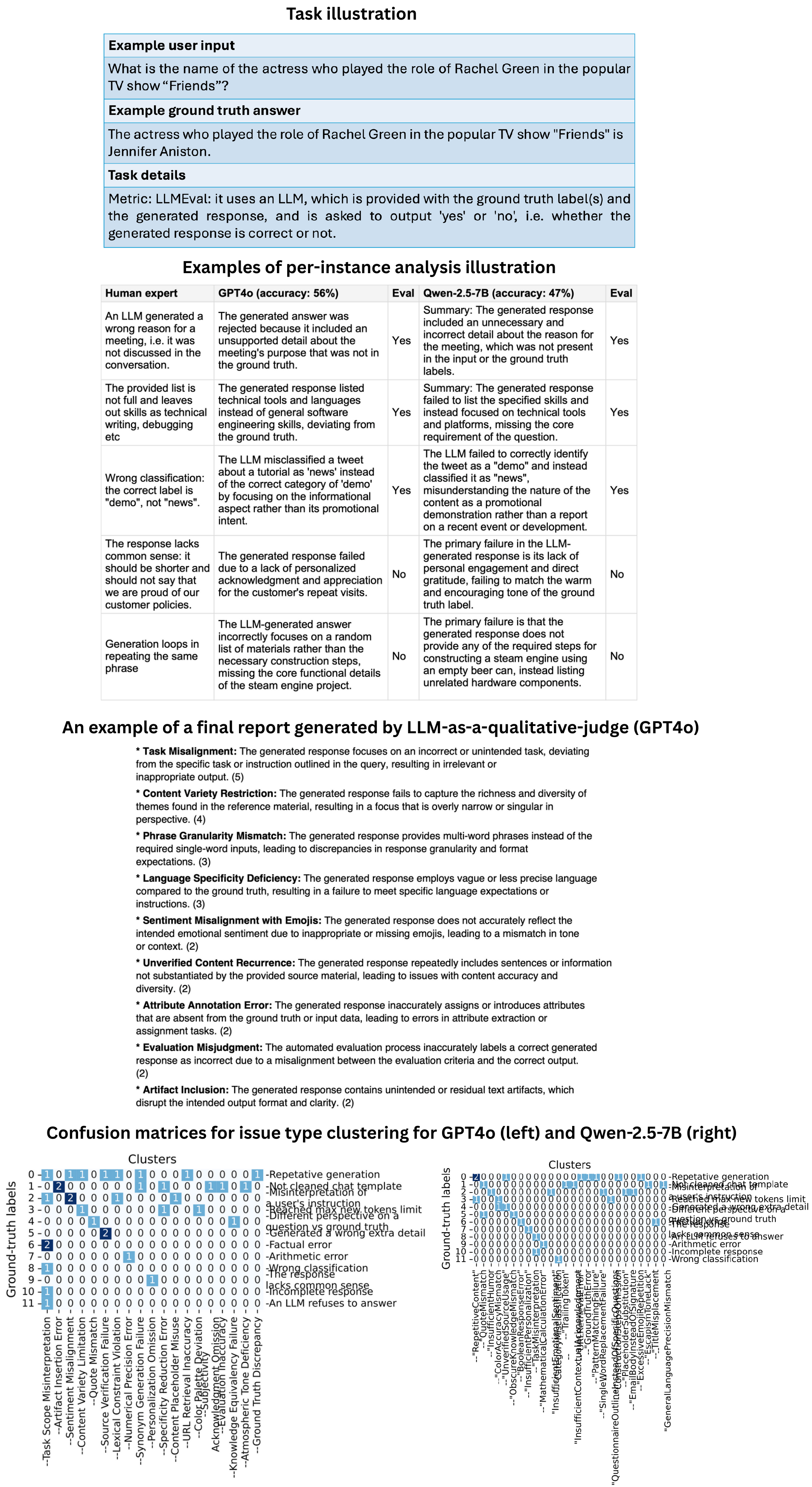}}

\subsection{Grade school math (GSM8K dataset)}
\center{\includegraphics[height=24cm]{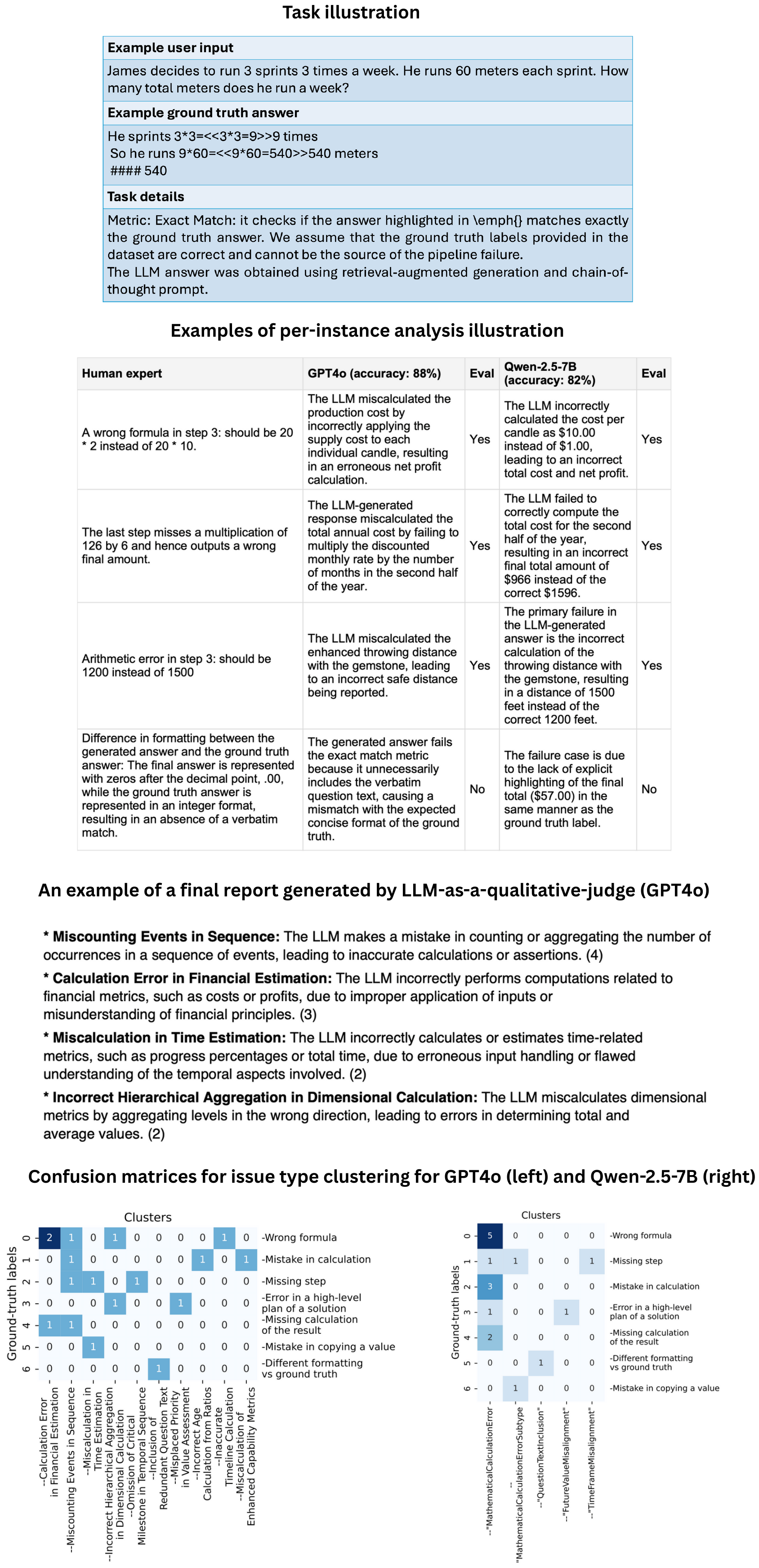}}

\subsection{Factoid QA in Russian (MKQA dataset)}
\center{\includegraphics[height=24cm]{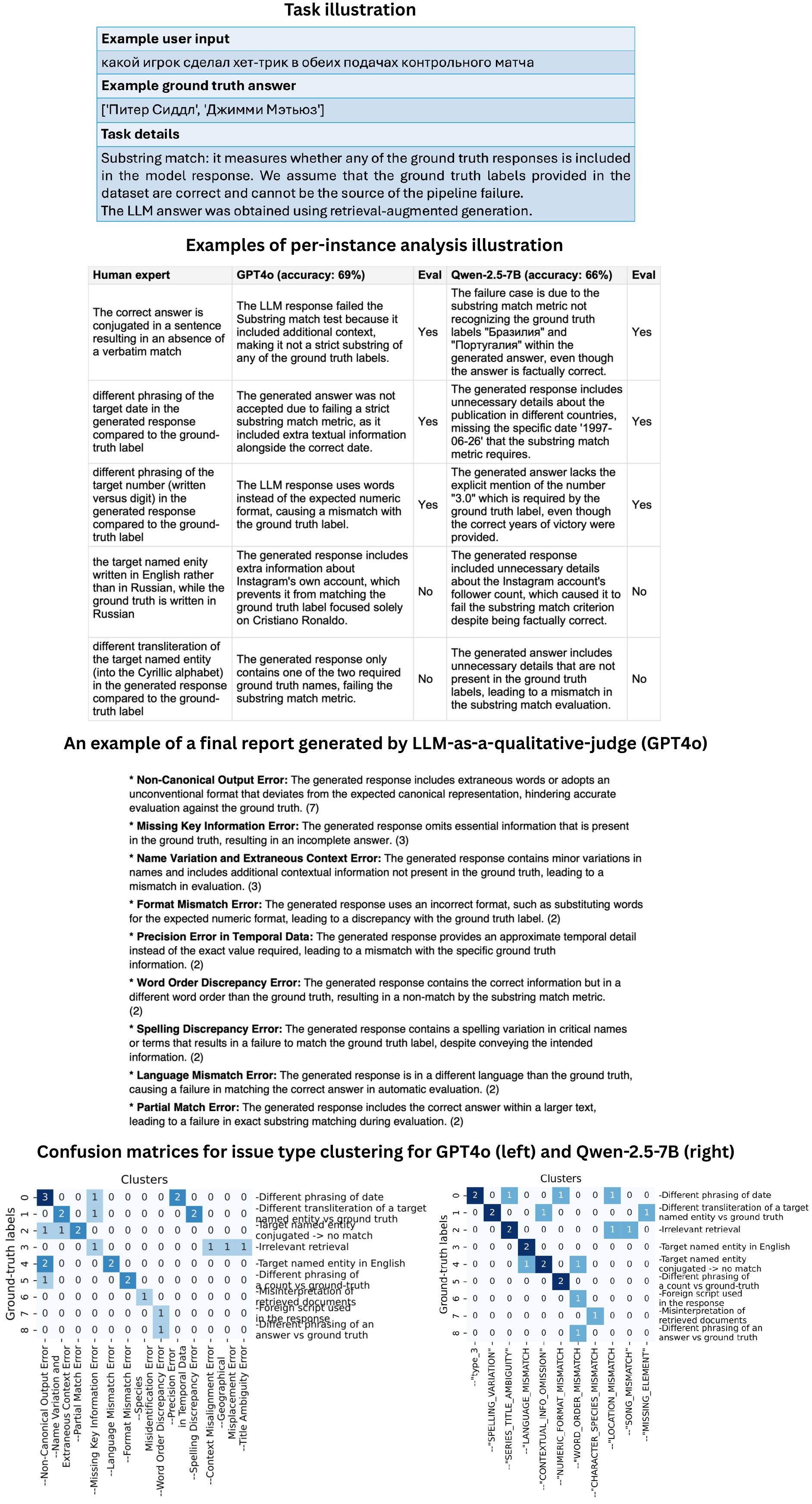}}

\subsection{Biomedical QA (BioASQ dataset)}
\center{\includegraphics[height=24cm]{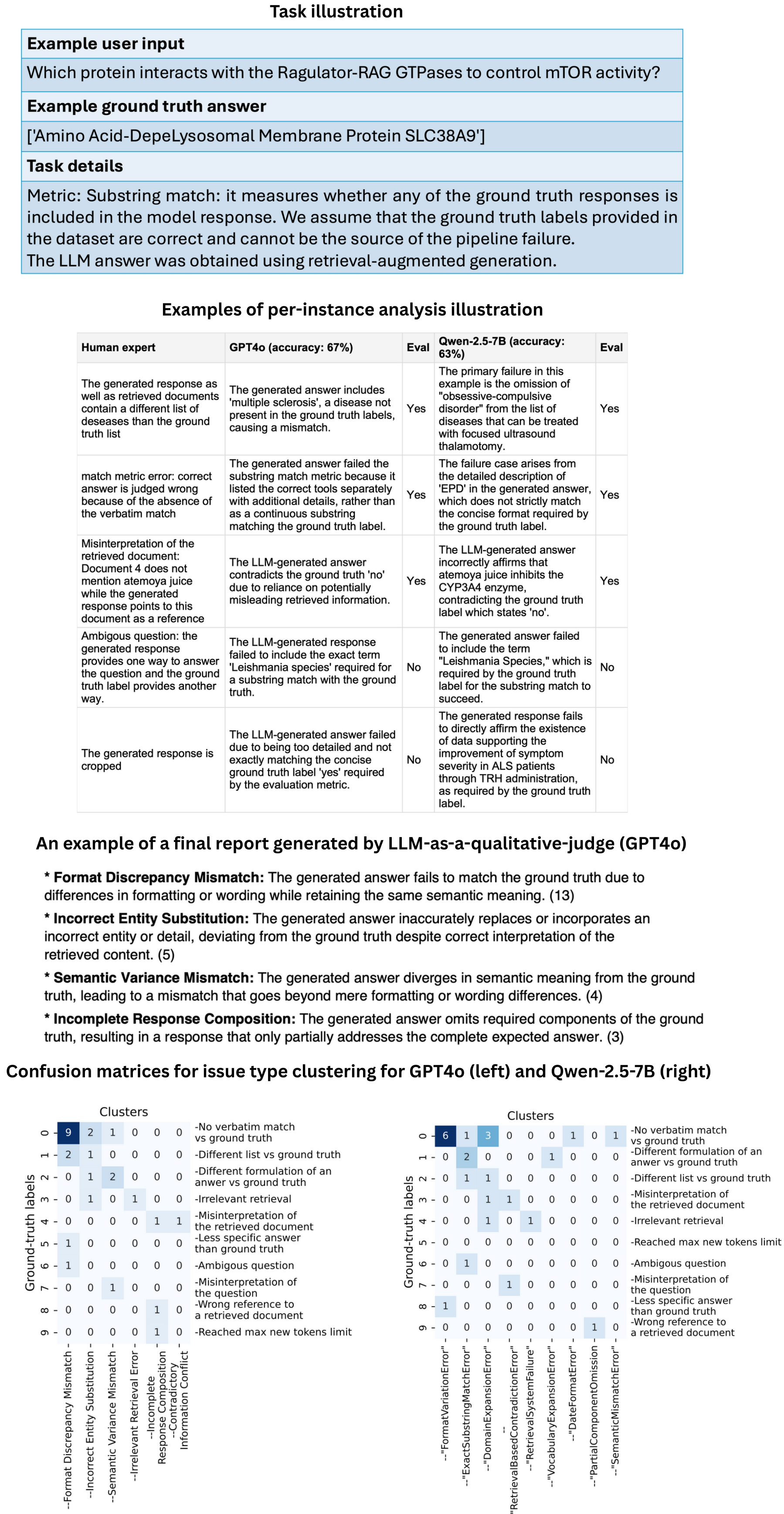}}

\subsection{Lifestyle forum QA (RobustQA dataset)}
\center{\includegraphics[height=24cm]{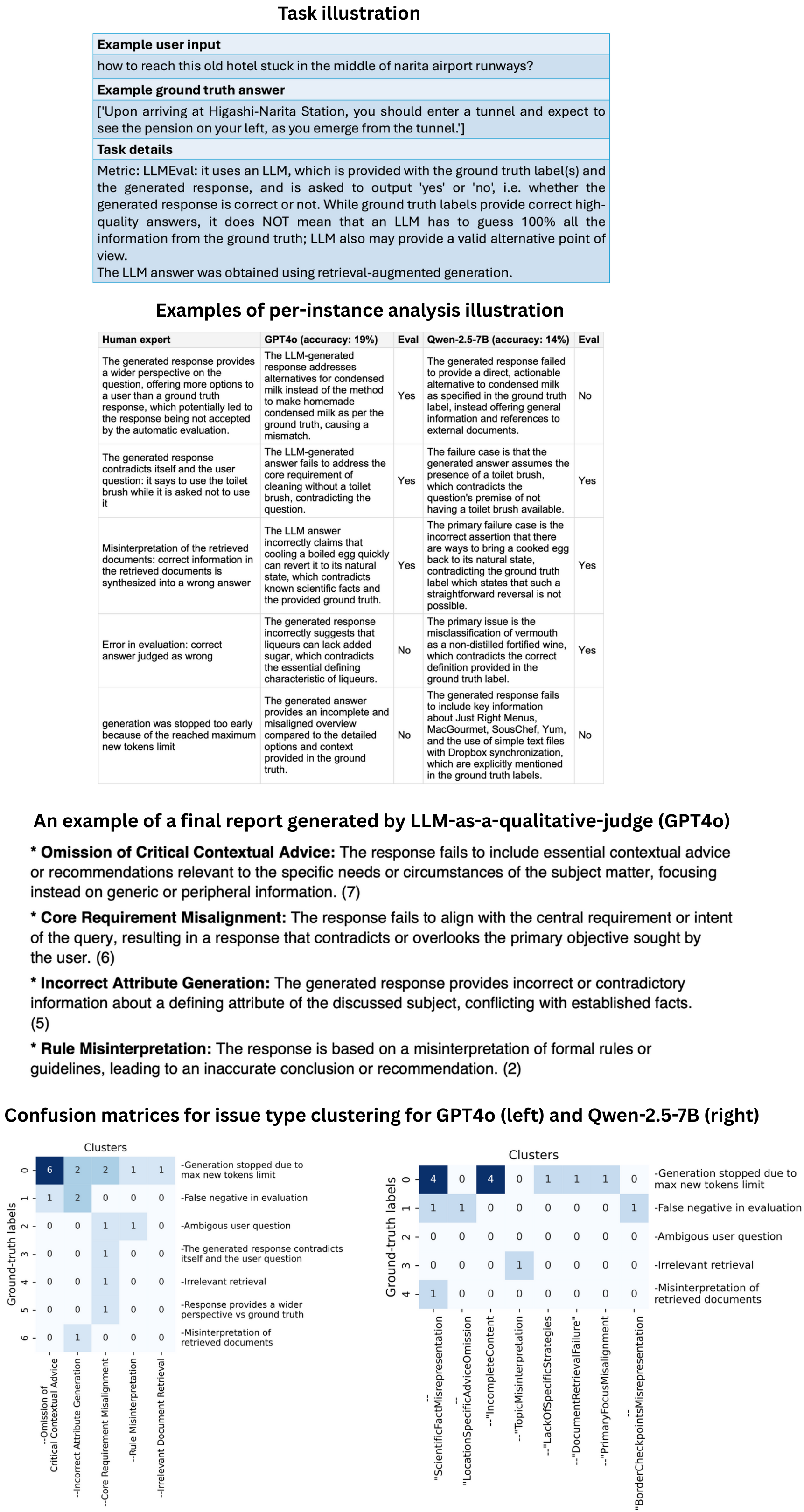}}

\subsection{Writing forum QA (RobustQA dataset)}
\center{\includegraphics[height=24cm]{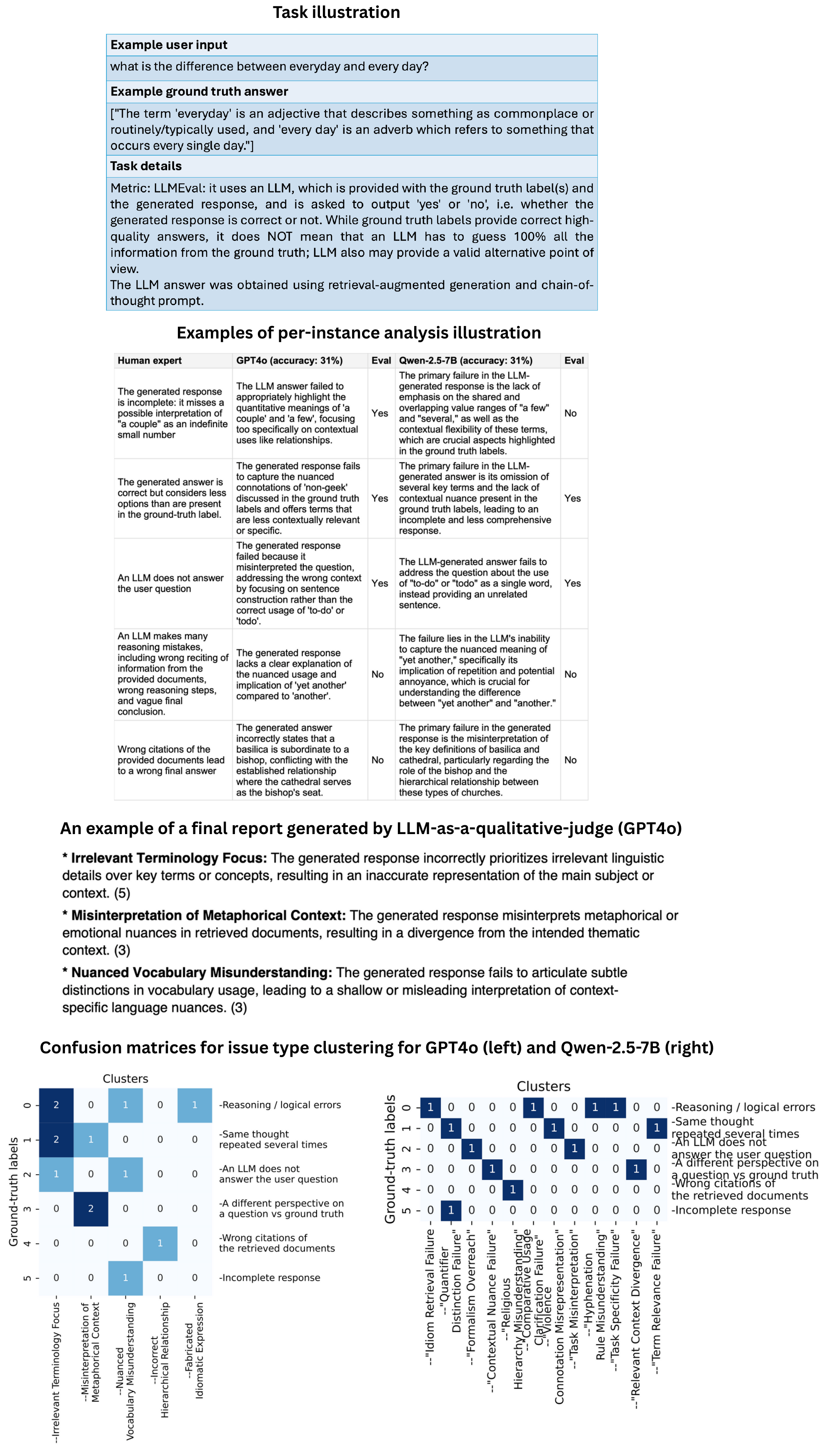}}

\subsection{Search engine queries (SearchQA dataset)}
\center{\includegraphics[height=24cm]{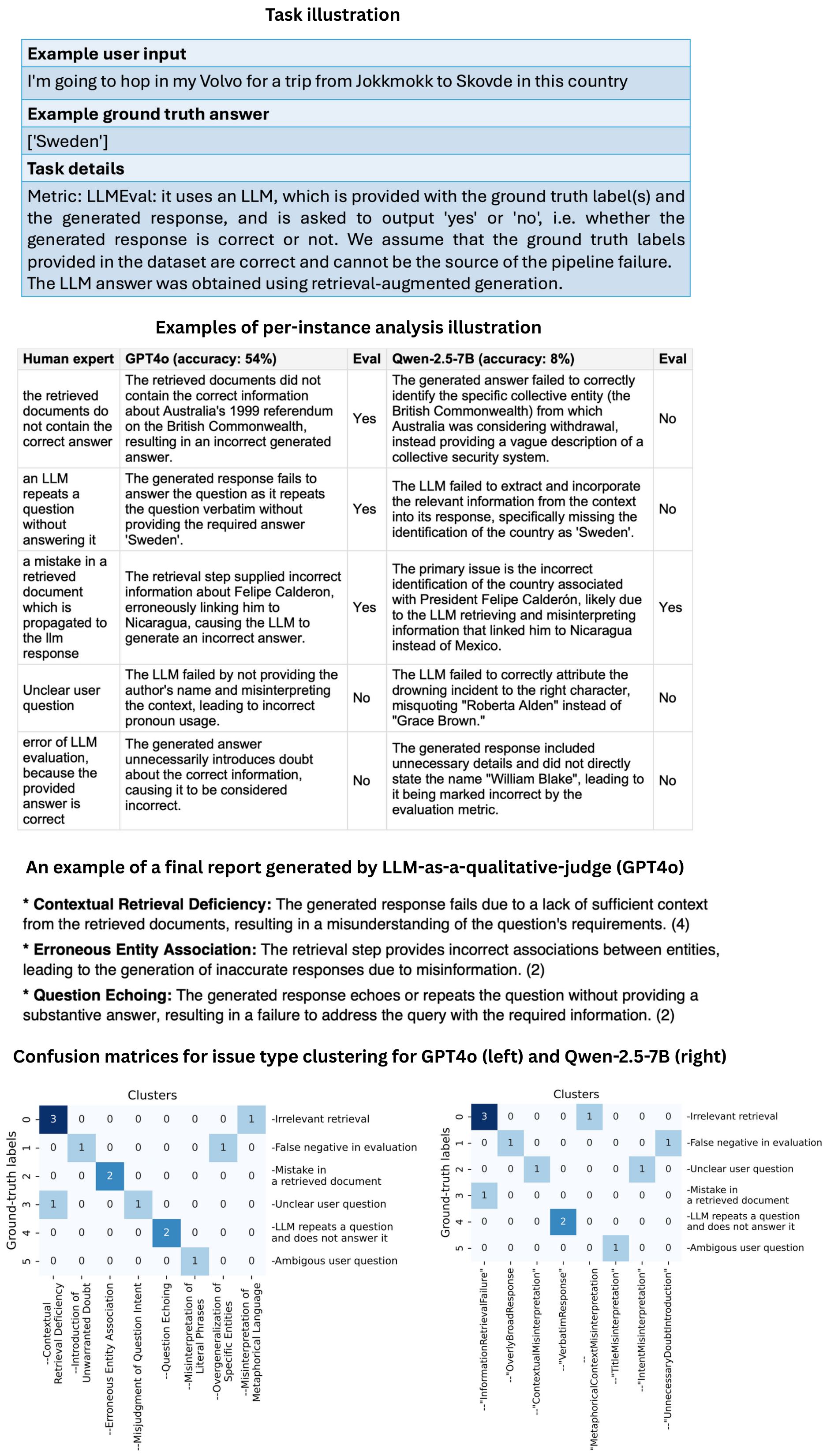}}

\subsection{Educational QA (SyllabusQA dataset)}
\center{\includegraphics[height=23cm]{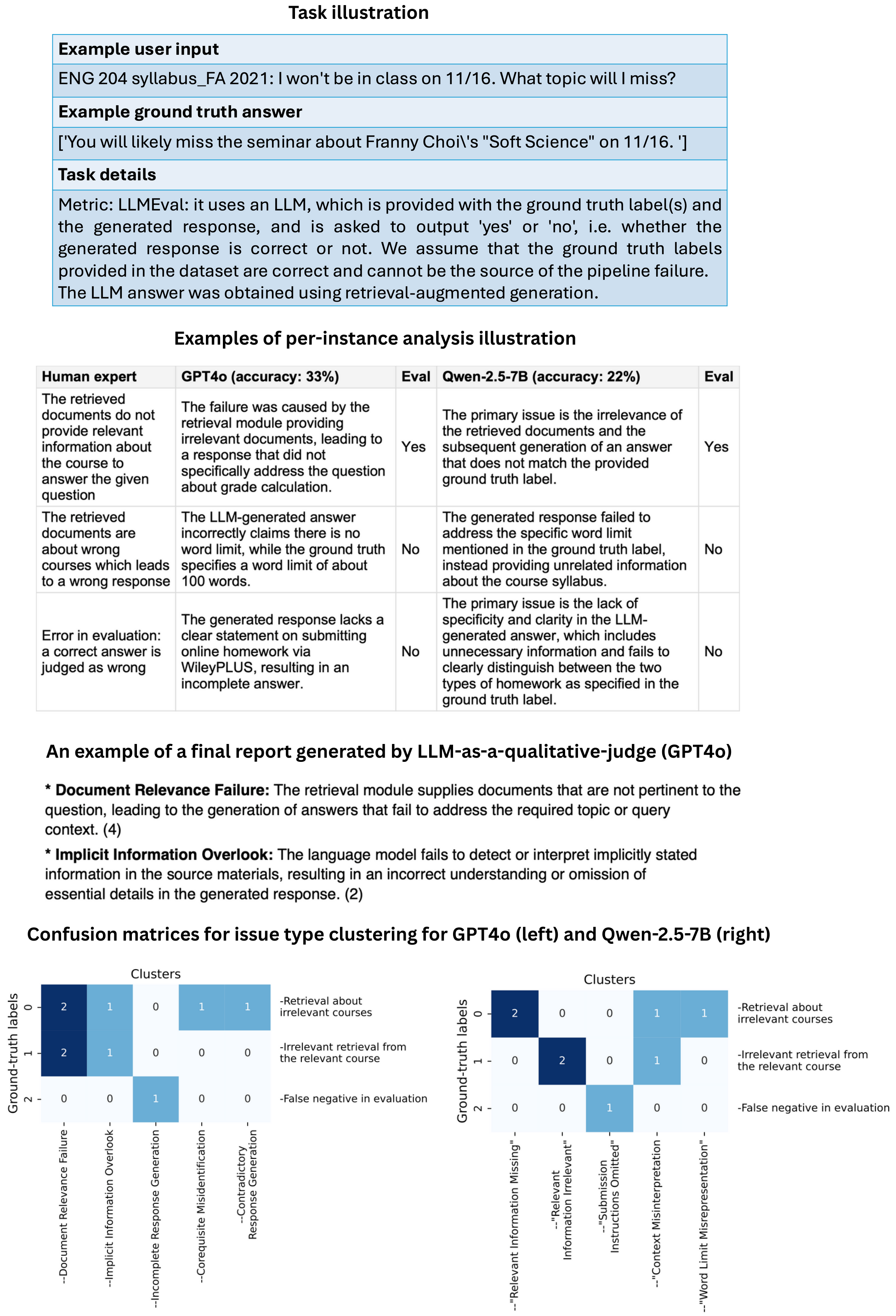}}

\twocolumn
\end{document}